\pdfoutput=1
\documentclass[acmsmall, manuscript]{acmart}

\AtBeginDocument{%
  \providecommand\BibTeX{{%
    \normalfont B\kern-0.5em{\scshape i\kern-0.25em b}\kern-0.8em\TeX}}}

\setcopyright{acmcopyright}
\copyrightyear{XXXX}
\acmYear{XXXX}
\acmDOI{XXXXXXX.XXXXXXX}

\acmJournal{TIIS}
\acmVolume{XX}
\acmNumber{X}
\acmArticle{000}
\acmMonth{00}

\graphicspath{{figures/}{pictures/}{images/}{./}} 

\usepackage{amsmath}
\usepackage{amsfonts}
\usepackage{tabu}                      
\usepackage{booktabs}                  
\usepackage{color}
\usepackage{xcolor}
\usepackage{paralist}
\usepackage{enumitem}  
\usepackage{algorithm}
\usepackage{algorithmic} 

\definecolor{commentout}{rgb}{0.5, 0.5, 0.5}
\newcommand{\commentclr}{\textcolor{commentout}}
\newcommand{\leftComp}{\textit{Image Space Exploration Component}}
\newcommand{\rightComp}{\textit{Neuron Space Exploration Component}}
\newcommand{\firstv}{\textit{Data Overview}}
\newcommand{\firstvScatter}{\textit{Scatterplot Views}}
\newcommand{\firstvScatterSingle}{\textit{Scatterplot View}}
\newcommand{\firstvMatrix}{\textit{Prediction Matrix View}}
\newcommand{\secondv}{\textit{Image Grid View}}
\newcommand{\thirdv}{\textit{Neuron Vulnerability View}}
\newcommand{\fourthv}{\textit{Receptive Field View}}
\newcommand{\fifthv}{\textit{Neuron Cluster View}}

\usepackage{multicol}
\begin{document}

\title{Visual Analytics of Neuron Vulnerability to Adversarial Attacks on Convolutional Neural Networks}

\author{Yiran Li}
\email{ranli@ucdavis.edu}
\affiliation{%
  \institution{University of California, Davis}
  \country{USA}
}
\author{Junpeng Wang}
\email{junpeng.wang.nk@gmail.com}
\affiliation{%
  \institution{Visa Research}
  \country{USA}
}
\author{Takanori Fujiwara}
\email{takanori.fujiwara@liu.se}
\affiliation{%
  \institution{Link\"{o}ping University}
  \country{Sweden}
}
\author{Kwan-Liu Ma}
\email{ma@cs.ucdavis.edu}
\affiliation{%
  \institution{University of California, Davis}
  \country{USA}
}

\begin{abstract}
    Adversarial attacks on a convolutional neural network (CNN)---injecting human-imperceptible perturbations into an input image---could fool a high-performance CNN into making incorrect predictions.
The success of adversarial attacks raises serious concerns about the robustness of CNNs, and prevents them from being used in safety-critical applications, such as medical diagnosis and autonomous driving.
Our work introduces a visual analytics approach to understanding adversarial attacks by answering two questions: (1) \textit{which neurons are more vulnerable to attacks} and (2) \textit{which image features do these vulnerable neurons capture during the prediction?}
For the first question, we introduce multiple perturbation-based measures to break down the attacking magnitude into individual CNN neurons and rank the neurons by their vulnerability levels.
For the second, we identify image features (e.g., cat ears) that highly stimulate a user-selected neuron to augment and validate the neuron's responsibility.  
Furthermore, we support an interactive exploration of a large number of neurons by aiding with hierarchical clustering based on the neurons' roles in the prediction.
To this end, a visual analytics system is designed to incorporate visual reasoning for interpreting adversarial attacks.
We validate the effectiveness of our system through multiple case studies as well as feedback from domain experts.

\end{abstract}







\maketitle

\section{Introduction} 

Recent advances in convolutional neural networks (CNNs) have achieved remarkable prediction performance in many applications,
including medical diagnosis~\cite{bakator2018deep,mei2020artificial}, object detection~\cite{ren2015faster}, and autonomous driving~\cite{grigorescu2020survey,tyrsa2017review}.
However, researchers have found that a subtle but intentional perturbation in a CNN's input image, called adversarial attacks~\cite{goodfellow2015explaining}, could cause a malfunction in a high-performance CNN. 
This phenomenon poses a severe threat to applications less tolerable to safety or security issues, such as road sign recognition of an autonomous driving system~\cite{eykholt2018robust}.

Due to adversarial attacks' critical impact on the safety and reliability of practical applications, research efforts have been devoted to developing defense techniques~\cite{papernot2016distillation, xu2018feature, madry2018towards,akhtar2018threat}. 
Understanding adversarial attacks' behaviors and unveiling CNN neurons' vulnerability is the key to effectively locating a CNN's weaknesses and defending the corresponding attacks~\cite{biggio2018wild}.

Pertinent to our work, multiple visual analytics approaches~\cite{das2020bluff, cao2021analyzing} identify critical neurons for adversarial attacks by comparing neurons' activation pathways on original and perturbed images. 
There are also approaches~\cite{hohman2019s, das2020bluff} that interpret each neuron's role in the prediction by examining image features (e.g., cat ears) that maximally activate a neuron or by synthesizing new interpretable image features~\cite{nguyen2016synthesizing, olah2017feature}. 
However, both groups of approaches have limitations.
The former group, i.e., extracting the activation pathways, is often very complicated, both logically and computationally, limiting the capability of being reproduced or scaled to large CNN architectures. 
The latter group of interpreting individual neurons' roles can only focus on reviewing a single neuron at a time. Additionally, the feature synthesizing method involves various types of randomness in its generation process, making it hard to compare neurons' roles and neurons' collective behaviors. 

Motivated by the above limitations in the existing works, we introduce an effective and intuitive method to interpret adversarial attacks from two aspects: (1) \textit{identifying more vulnerable neurons to attacks} and (2) \textit{understanding image features that the vulnerable neurons capture during the prediction.}
For the first aspect, we apply data perturbations to both input images and neurons and then sort the neurons with multiple measures that we design to identify the most vulnerable neurons.
For the second, we utilize the receptive fields (RFs)~\cite{long2014do}, which reveal an input image's area that greatly stimulates each neuron, and visually compare the RFs for a set of images before and after the perturbations. 
In addition to the analysis of a single neuron, to efficiently explore a large number of neurons and interpret their semantics, we support the exploration of a group of neurons clustered in a hierarchy based on their roles in the prediction.

We further integrate these functionalities into a coordinated visual analytics system to allow our users, i.e., domain experts using CNNs and concerned about adversarial attacks, to understand their adversaries through interactive exploration of the two aspects for multiple granularities of input images and neurons.
Specifically, our system first facilitates the exploration of the input images and their attacked counterparts (i.e., the \textit{image space}) by providing an overview of the images and sorting them by their perturbation magnitude or prediction probability changes. Second, focusing on an image of interest, our system helps domain experts efficiently explore a large number of the CNN's neurons (i.e., the \textit{neuron space}) by sorting them based on their vulnerability and clustering them based on their role similarity. Lastly, the coordination among multiple views of our system empowers effective analysis across the two spaces and leads to meaningful interpretations of different attacking scenarios. 
Using the visualization system, domain experts can obtain an overview of the entire image and neuron spaces and dive into the detailed analysis of specific images or neurons per their selections following our ranking approach. Through the efficient multi-level exploration within and across the two spaces, domain experts can identify and reason about the weaknesses of the CNN against adversarial attacks.
We conduct case studies together with domain experts to validate the findings and the derived insights.

In short, the main contributions of our work are as follows:
\begin{compactitem}
    \item We introduce multiple measures to rank neurons by their vulnerability to attacks.
    \item We facilitate the exploration and comparison of an excessive number of neurons in a CNN with a hierarchy that is constructed based on the individual neurons' roles in the predictions. 
    \item We design an integrated visual analytics system supporting flexible explorations across multiple input images and neurons to help domain experts gain actionable insights from the interpretation of adversarial attacks.
\end{compactitem}

\section{Related Work}
Our work addresses the interpretation of adversarial attacks on CNNs, which falls under the research for explainable machine learning (ML). 
We highlight representative research on the related topics.

\textbf{\textit{Visualizations for Explainable Machine Learning.}} 
A rich amount of visualization solutions have been introduced recently for the understanding, debugging, and refinement of ML models~\cite{choo2018visual}. 
These visualizations cover traditional ML models (e.g., ~\cite{ma2019explaining, ma2020visual}), ensemble learning models (e.g.,~\cite{liu2017visual,wang2021investigating,zhao2018iforest}), and deep learning models (e.g.,~\cite{cao2021analyzing, das2020bluff, wang2018dqnviz}). They have successfully proven that visualization can help machine learning models in multiple aspects, e.g., transparency, reliability, fairness, and robustness~\cite{cao2021analyzing, das2020bluff,ma2019explaining}. The robustness of deep CNNs (when being attacked by adversarial examples~\cite{goodfellow2015explaining,madry2018towards}) is the focus of our work.

\textbf{\textit{Visual Analytics of Adversarial Attacks.}} Adversarial attacks~\cite{goodfellow2015explaining} draw increasing research attention, and so do their interpretations. From the visualization field, we identified three closely related works. 
First, AEVis~\cite{cao2021analyzing} extracts the data paths of a \textit{benign} image (i.e., non-attacked image) and its \textit{adversarial} counterpart (i.e., attacked image) from a CNN and investigates diverging and merging patterns of the paths to know where the attacks happened in the CNN. 
Second, Bluff~\cite{das2020bluff} separates CNN neurons into four groups, i.e., neurons that activated only for the benign class, only for the adversarial class, for both, and only for the attacked images. These four groups of neurons are connected to form a graph, from which one can infer what original features were kept and what features were newly injected during the attack. 
Third, Ma et al.'s visual analytics system~\cite{ma2019explaining} is designed to examine data poisoning attacks on conventional regression models while categorizing the attacking process into four different levels, i.e., instance, model, feature, and local structure levels.
These works, however, come with several limitations. The data path extraction methods in AEVis~\cite{cao2021analyzing} and Bluff~\cite{das2020bluff} not only rely on certain assumptions of CNNs' behaviors but also involve the intricate extraction process that may lead to volatile interpretations due to the randomness in the computation. The work of Ma et al.~\cite{ma2019explaining} addresses the attacks on simple regression models, and their solution cannot be easily extended to complex CNNs, where investigating an enormous number of neurons is equally (or even more) important as (than) the analyses of input data and their features.

Our work develops a visual analytics approach that compensates these limitations and provides new ways to understand adversarial attacks.
Instead of extracting critical neurons from the data paths of benign and adversarial images, we focus only on the last convolutional layer and provide an exhaustive exploration of all neurons in this layer.
By focusing on the last layer, our approach does not depend on any assumptions on the internal model structure and is thus less model-specific and adaptable to any type of CNNs.
Also, we can analyze adversarial attacks with logically intuitive perturbation-based measures.
With these measures, we can order and identify vulnerable neurons from diverse perspectives.
Furthermore, by exploring all the neurons organized by the similarities of their roles in predictions, we can understand the attacks from broader perspectives while avoiding false insights due to the neurons with duplicate roles.

\textbf{\textit{CNN Neuron/Filter Interpretation.}}
In the field of computer vision, multiple feature visualization techniques have been introduced to disclose image features that individual CNN neurons/filters have captured. They can be categorized into two groups based on the computation order employed in the techniques. The \textit{forward techniques} use the forward-propagation of CNNs to identify input pixels (local features) that excite a neuron the most. The RF~\cite{long2014do} is a canonical example, which we explain in detail in \autoref{sec:receptive_field}. 
The discrepancy map~\cite{zhou2015object} is another example in this category, which perturbs individual pixels of an input image and then refers to the neurons' active-level change caused by the perturbations in order to locate the sensitive features (i.e., groups of pixels). 
The \textit{backward techniques} rely on the backward-propagation of CNNs to synthesize image features that maximally excite a neuron. For example, activation maximization~\cite{nguyen2016synthesizing} keeps CNNs' trained parameters untouched but synthesizes the input image to maximize a neuron's activation through gradient ascent. DeepDream~\cite{mordvintsev2015deepdream} extends activation maximization by initiating the synthesized image with an image to be analyzed to preserve the spatial correlations of image features. Feature visualization~\cite{olah2017feature} is also based on activation maximization but enhances it by improving the diversity of the synthesized features. Our work employs a forward technique (specifically, RF extraction~\cite{long2014do}) to visualize features captured by individual CNN neurons/filters as it is usually more computationally efficient and the derived features (i.e., patches of the raw input images) are more intuitive to understand. 

\section{Background and Motivation}
\label{sec:background}
This section provides a general overview of adversarial attacks, CNNs, and the extraction process of the RF of a neuron. Also, we elaborate on the necessity of visualizations for the interpretations of adversarial attacks, motivating our work.   

\subsection{Adversarial Attacks on CNNs}
\label{sec:adv_attack}

Adversarial attacks on CNNs~\cite{akhtar2018threat,goodfellow2015explaining} add small imperceptible perturbations to an image to fool the CNNs into incorrectly predicting the image's class. The process can be formulated as follows. Given an image $x$ and its true label $y$ ($y = \{1, \cdots, c\}$ where $c$ is the number of classes), a CNN denoted as a function $f()$ takes $x$ as input and generates a probability distribution across all labels: $f(x) = [p_1, p_2, \cdots, p_c]$.
If the CNN is well-trained and can correctly predict this instance, $argmax_{i}(p_i)=y$.
Adversarial attacks add a subtle perturbation $\sigma$ to $x$, resulting in a new image $x'$, such that $f(x+\sigma) = [p_1', p_2', \cdots, p_c']$ and $argmax_{i}(p_i') \neq y$. $argmax_{i}(p_i)$ and $argmax_{i}(p_i')$ are called \textit{benign and adversarial labels}, respectively. 
Similarly, non-attacked images and attacked images are often called \textit{benign and adversarial images}, respectively.
The perturbation $\sigma$ is often very small such that a benign image $x$ and its adversarial counterpart $x'$ are visually identical.

In general, implementations of adversarial attacks can be categorized into white-box and black-box attacks~\cite{akhtar2018threat} according to whether the attacking process needs to know the CNNs' internal details or not. 
This paper adopts a white-box approach using projected gradient descent (PGD)~\cite{madry2018towards}, which perturbs the input image with a given perturbation budget so that the cross-entropy loss between the predicted probability and its true label is \textit{maximized}.
More specifically, PGD iteratively adjusts a perturbation, $\sigma$, such that each iteration moves the corresponding probability distribution, $f(x + \sigma)$, toward the gradient \textit{ascent} direction of the loss function while restricting $\sigma$ with a condition, $\lVert \sigma \rVert_\infty \leq \epsilon$, where $\epsilon$ is a perturbation budget.
Besides its popularity and efficiency, PGD is proven the strongest attack that utilizes the first-order information of the network~\cite{madry2018towards}; thus, we select PGD as an adversarial attacking method.
However, we should note that our visual analytics approach is agnostic on attacking methods, and can be easily adapted to other methods. 

Although researchers are well aware of adversarial attacks, there exist only a few clear and easily accessible interpretations of the attacks. 
Many questions regarding why the attacks could fool CNNs remain unanswered, such as: What kind of image features are more vulnerable to an attack? 
How are the roles of neurons changed by an attack? 
The answers to these questions need substantial visual evidence, motivating us to introduce novel visual reasoning techniques and develop a visual analytics system for coordinated analyses.

\subsection{CNN and Neurons' Receptive Field}
\label{sec:receptive_field}
\begin{figure}[b]
 \centering
 \includegraphics[width=0.8\columnwidth]{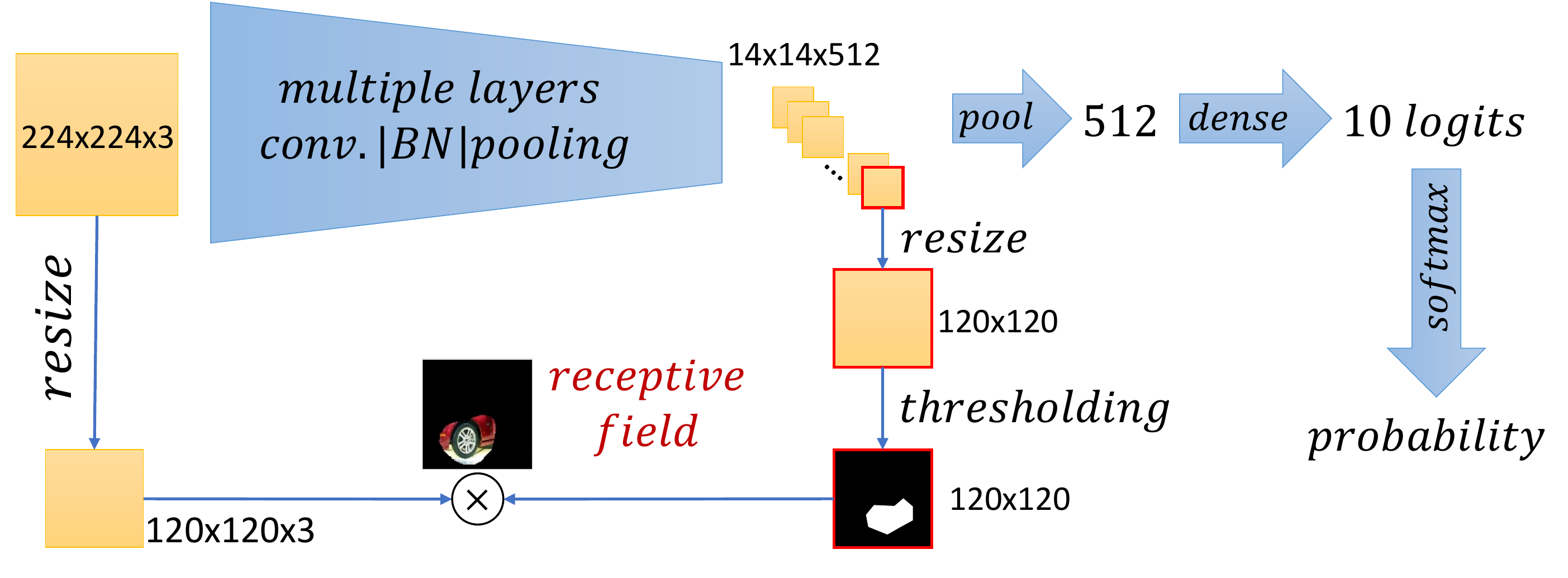}
 \caption{A typical CNN architecture (in blue) and the RF computation for a neuron from the last convolutional layer.}
 \label{fig:receptive}
\end{figure}

CNNs are the most powerful image classification model thus far. 
It chains a sequence of convolutional layers, usually with batch normalization (BN) and pooling layers in between, to extract image features. These features are then combined through dense layers to generate the final class probability (see the blue parts in Fig.~\ref{fig:receptive}). The outputs from each convolutional layer are features extracted in multiple channels (or image patches). Each of the outputs is called a feature map. For example, in Fig.~\ref{fig:receptive}, the extracted feature maps from the last convolutional layer are of shape $14{\times}14$ and we have 512 image patches, each of which reflects the extracted output from the corresponding convolutional filter. Note that ``neuron'' and ``filter'' are used interchangeably in this paper as with existing works.

The term, receptive field (RF), originates from neuroscience, which refers to the area of a body surface where a stimulus could elicit a reflex~\cite{alonso2009receptive}. 
Extended to the context of CNNs, RF refers to the regions of the inputs that affect a particular unit the most~\cite{goodfellow2016deeplearning}.
Therefore, the semantics reflected by a set of pixels in the input image is often used to interpret the neuron. 
Zhou et al.~\cite{zhou2015object} implemented and compared two methods of computing and visualizing RFs.
The first method is the theoretical approximation, which uses the feature map of the corresponding neuron to perform image segmentation based on the selected threshold.
The second method empirically captures the RFs with estimations of the discrepancy map.
Although the second method produces close-to-exact RFs, the computation is much more expensive as the discrepancy map calculation requires forward passes for the image copies with noise patches at all possible locations.

In our work, we use the first method because it is efficient and also sufficient to produce reasonable and meaningful results.
We describe the details of this method in the following four steps while referring to the example in \autoref{fig:receptive}:
\begin{compactenum}
    \item Resize the input RGB image to the desired RF size (e.g., from $224{\times}224{\times}3$ to $120{\times}120{\times}3$).
    
    \item For the target neuron at the desired convolutional layer (e.g., the last convolutional layer in \autoref{fig:receptive}), resize its feature map to the desired RF size (e.g., from $14{\times}14$ to $120{\times}120$).
    
    \item Specify a threshold value to binarize the enlarged feature map from the second step.
    
    \item Apply elementwise multiplication between each of the RGB channels of the resized input and the binarized feature map. 
\end{compactenum}
In the third step, the feature map is binarized so that it becomes a mask upon the image in the fourth step, where only the RF in the image is revealed and highlighted.
The threshold value used in the third step controls the number of pixels uncovered.
Note that the threshold can be interactively adjusted, as described in~\autoref{sec:neuron_space}.
The output from the last step is the RF. 
For example, the RF result in \autoref{fig:receptive} reveals that the target neuron extracts the wheel region of the input image from a \texttt{car} class. 
If we can see the same result across a collection of \texttt{car} images, we can infer that the image feature representing wheels always excites this neuron.

In this work, we focus only on neurons in the last convolutional layer of a CNN to compute the RFs because of the following two reasons.  
First, these neurons' RFs are more semantically interpretable than those from the other convolutional layers. 
As has been crystallized~\cite{zeiler2014visualizing, hohman2019s}, CNN neurons in lower layers (i.e., layers closer to the input) extract basic color and shape primitives (e.g., line or curve patterns), whereas higher layers' neurons combine previous neurons' roles to extract object-level features that are more related to semantics (e.g., cat head and car wheel). 
Focusing on the top level of this learning hierarchy can thus better show the semantics.  
Second, neurons in the last convolutional layer are right before a dense layer that linearly combines the extracted features to derive the prediction probability. 
These neurons have a more direct impact on the final prediction probability, which helps us understand the prediction probability changes caused by adversarial attacks.

On the one hand, a convolutional layer could have hundreds of neurons, resulting in hundreds of RFs for a single input image (\textit{one image, many neurons}). 
On the other hand, the RF analysis of a single neuron also needs the support of numerous input images to extract the shared semantics and to build confidence during interpretation (\textit{one neuron, many images}). This brings the need for easy exploration across a space of input images (we call \textit{image space}) and a space of neurons (\textit{neuron space}) to identify images/neurons of interest and interpret adversarial attacks from both aspects. 

\section{Requirement Analysis}
Based on the stated research questions, the need for visualization solutions discussed in \autoref{sec:background}, and the review of the closely related literature~\cite{cao2021analyzing,das2020bluff}, we extract several common analysis requirements of our target users---ML experts with a demanding need to interpret adversarial attacks. 
We have distilled and iteratively refined the following requirements for our visual analytics approach and system.

\setdefaultleftmargin{1em}{1em}{}{}{}{}
\begin{compactitem}

\item \textbf{R1:}
\textbf{Multi-level exploration of the \textit{image space}}. As images generated through adversarial attack experiments usually come out in large numbers and multiple classes, our first requirement is to explore the image space and locate images of interest. This requires our system to:

\begin{compactitem}
\item \textbf{R1.1:} provide an overview of benign images, adversarial images, and their distributions across the predictions;
\item \textbf{R1.2:} filter images based on their feature similarity (e.g., similar \texttt{cat} images) or prediction similarity (e.g., all \texttt{cat} images being mispredicted into \texttt{dog});
\item \textbf{R1.3:} prioritize and select images based on attack-related measures, e.g., perturbation magnitude or prediction changes.
\end{compactitem}

\item \textbf{R2:} \textbf{Efficient exploration of \textit{the neuron space.}} 
Even within a single layer, CNNs usually have an excessive number of neurons; consequently, the exploration of the neuron space becomes nontrivial. 
To achieve efficient exploration, we should address questions of which neuron we should explore, how we can locate it, and how we can concurrently explore many neurons.
Specifically, our system needs to be able to:

\begin{compactitem}
\item \textbf{R2.1}: effectively identify the most vulnerable neurons to attacks and prioritize them for analysis;
\item \textbf{R2.2}: intuitively reveal the role of a neuron in the prediction and how its role is changed by adversarial attacks;
\item \textbf{R2.3}: collectively explore neurons that have similar behaviors (i.e., having similar responded image features) before and after adversarial attacks to improve the exploration efficiency.
\end{compactitem}

\item \textbf{R3: Interpretation of adversarial attacks \textit{across the image and neuron spaces}.}
Frequently, the image and neuron spaces need to be explored concurrently to interpret adversarial attacks as the image space carries meaningful semantics and the neuron space encodes the CNN prediction mechanisms. 
Only when bridging the two spaces can we externalize how a CNN works through human-understandable semantics. The cross-space analysis can focus on a one-to-one relationship (i.e., how a neuron works on an image), a one-to-many relationship (i.e., how a neuron works on a collection of similar images), or a many-to-one relationship (i.e., how different neurons behave to a focused image). 

\end{compactitem}

\section{Visual Analytics Methodology}

\begin{figure*}[tbh]
 \centering
 \includegraphics[width=\textwidth]{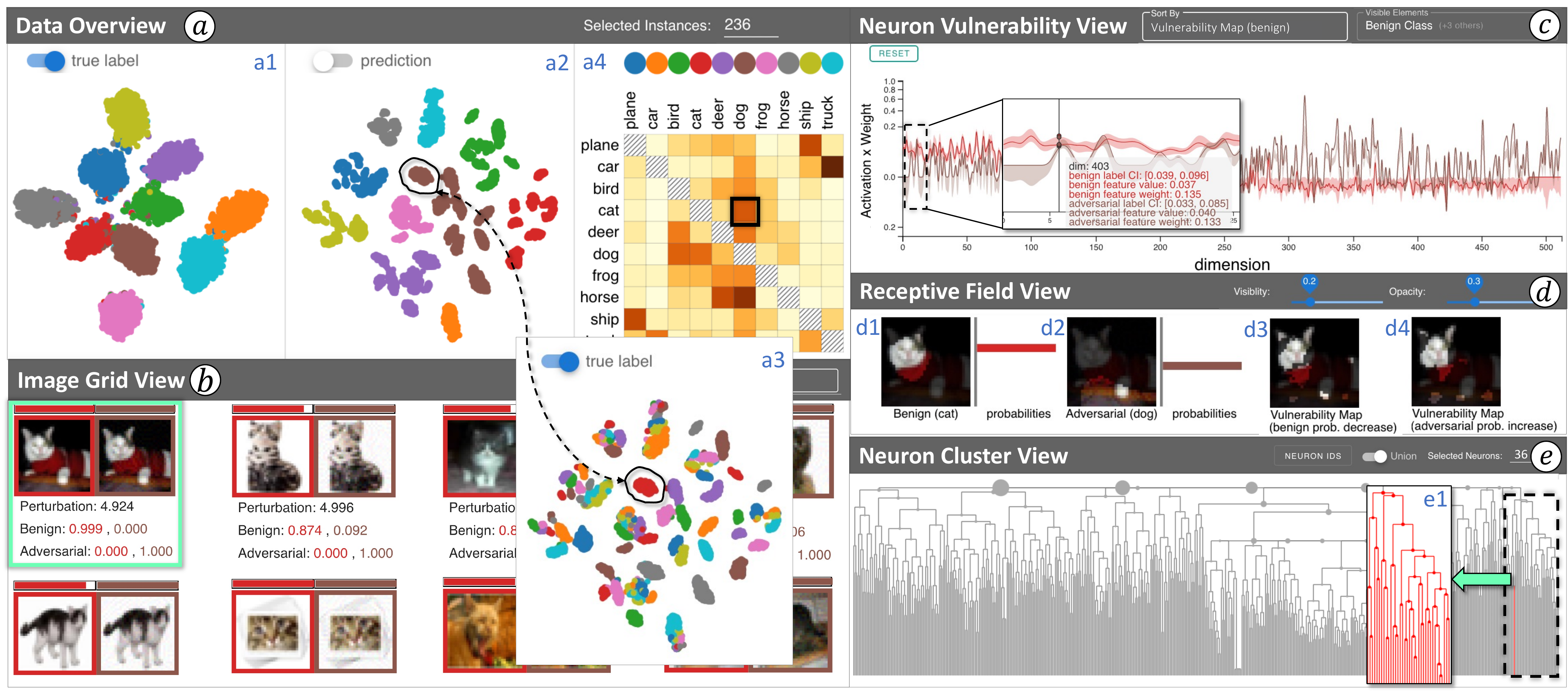}
 \caption{Using our visual analytics system to explore and interpret the adversarial attacks. The \firstv{} (a) visualizes the benign images' similarities (a1) and the adversarial images' similarities (a2) based on the CNN penultimate layer’s activation. The color encoding of the points can be switched between using true labels and using predicted labels (a2 and a3).
 The Prediction Matrix (a4) shows the distribution of images based on true and adversarial labels, where the darker orange color represents more images in the cell. 
 The \secondv{} (b) lists benign images and adversarial counterparts corresponding to the selection made in the \firstv{} (here the \texttt{cat}-\texttt{dog} cell is selected in a4). 
 For the selected image in the \secondv{}, the \thirdv{} (c) organizes neurons based on the differences in their values of activation${\times}$weight. 
 The \fourthv{} (d) informs the RFs and vulnerability maps to help interpret each neuron's behavior. 
 The \fifthv{} (e) shows the dendrogram constructed based on the similarity of each neuron's behavior and facilitates efficient interpretation of multiple neurons' behaviors. 
}
\label{fig:system}
\end{figure*}

This section describes our visual analytics approach and system.
As shown in \autoref{fig:system}, the system consists of five main views.
The \leftComp{} at the left-hand side of the system provides the \firstv{} (\autoref{fig:system}-a) and \secondv{} (\autoref{fig:system}-b) to support \textbf{R1}.
The \rightComp{} at the right-hand side comprises the \thirdv{}, \fourthv{}, and \fifthv{} (\autoref{fig:system}-c, d, e, respectively) to support \textbf{R2}. 
All five views are coordinated, enabling users to flexibly switch between the image space and neuron space to connect them for the cross-space explorations (\textbf{R3}).

\subsection{Image Space Exploration}
Our analysis focuses on test-set images that have true positive predictions before adversarial attacks. We use PGD to attack these images, and only select the successfully attacked ones for our analysis. Each of the original/benign images also comes with an attacked/adversarial counterpart. The \leftComp{} is designed to explore these images, targeting two goals: (1) grasping an overview of the benign images, the adversarial images, and their relations regarding the prediction outcomes from the studied CNN; (2) exploring a subset of images of interest (benign-adversarial pairs) to identify significant/representative ones to be further analyzed.

The \firstv{} (\autoref{fig:system}-a) is designed for the first goal. The two \firstvScatter{} lay out all benign (\autoref{fig:system}-a1) and adversarial (\autoref{fig:system}-a2) images. The layouts are based on t-SNE projections~\cite{maaten2008tsne} using the CNN penultimate layer's activation (e.g., 512 features produced by a pooling layer in \autoref{fig:receptive}) of individual images as input. The points are colored by either true labels or predicted labels based on the user's selection. For example, in \autoref{fig:system}-a2 and a3, the predicted and true labels are used, respectively. 
The \firstvMatrix{} (\autoref{fig:system}-a4) visualizes a heatmap that presents the distribution of all images, using their true label as the row index and the adversarial label (i.e., the predicted label of the adversarial image) as the column index. For example, if a \texttt{cat} image is mispredicted as a \texttt{dog} after being adversarially perturbed, it falls into the \texttt{cat}-\texttt{dog} cell, which is annotated in \autoref{fig:system}-a4. The darkness of each cell's orange color represents the number of instances falling into the cell.
Users can select a subset of images to perform detailed analyses with the other views by lasso-selecting points in either of the two \firstvScatter{} or clicking a cell of the \textit{Prediction Matrix}.

The \secondv{} (\autoref{fig:system}-b) is designed to meet the second goal. 
The selected images in the \firstv{} are presented as a grid of image pairs (left: benign image, right: adversarial counterpart).
\autoref{fig:system} shows a result after selecting the \texttt{cat}-\texttt{dog} cell from the \firstvMatrix{}.
The border color of each image reflects the predicted label and the length of the filled color bar above each image shows the predicted probability for the benign and adversarial labels, respectively.
To help users identify the interesting image pairs, this view supports sorting of the image pairs with {four} measures: the increasing order of perturbation magnitudes (i.e., the $L_2$ distance between the benign and adversarial images), the decreasing order of perturbation magnitudes, the decreasing order of the benign label's probabilities, or the increasing order of the adversarial label's probabilities.
As stated in \autoref{sec:adv_attack}, the adversarial attack aims to make a mismatch between benign and adversarial labels. 
This can be caused by decreasing the benign label's probability, increasing the adversarial label's probability, or both. 
Also, the perturbation magnitude represents the general strength of the attack.
Thus, the four measures above are useful to order the images by how intensely/successfully the adversarial attacks disturb the prediction. 
The exact perturbation magnitude and the probabilities of being benign and adversarial classes are listed below each image pair as a reference.

With the above views, the \leftComp{} allows users to explore the image space in a top-down manner with an ``Overview+Details'' strategy. The selected image pairs are used in the next component for further analysis.

\subsection{Neuron Space Exploration}
\label{sec:neuron_space}

Focusing on the selected pair of images, we extract the information of neurons in the last convolutional layer from the studied CNN and provide multiple ways to explore them. 
This section consolidates the exploration and interpretation of a large number of neurons into two questions and explains our way of answering them. 

\subsubsection{How to Identify the Most Vulnerable Neurons?}
\label{sec:newmetrics}
A straightforward answer is to sort the neurons based on the difference between neuron activations for the benign and adversarial images. 
However, neuron activations do not directly encode the image semantics, which is important for the understanding of adversarial attacks.
To systematically evaluate neuron vulnerabilities and associate vulnerable neurons with the image semantics, we develop a more robust sorting by introducing two new measures that quantify the vulnerability through the perturbation of the input images or the output neurons. Perturbation-based methods are widely used for ML analysis and interpretation, which characterize how controlled local changes in the input affect the model behavior~\cite{liu2019nlize, greydanus2018visualizing}. 

\setlength{\textfloatsep}{0.3cm}
\begin{algorithm}[t]\footnotesize
\caption{Generate vulnerability maps by region substitution.}
\label{alg:vulmap}
\begin{algorithmic}[1]
\STATE $w, h, k, s, c$\ \commentclr{// image width, height, region size, stride, number of classes}
\STATE $bImg, aImg$\ \commentclr{// benign image, adversarial image}
\STATE $ bId, aId$\ \commentclr{// benign class id, adversarial class id}
\STATE $[y_1, y_2, ..., y_c] = CNN(bImg)$
\FOR{$i=0;\ \ i < w;\ \ i{+}{=}s\ \ $}
\FOR{$j=0;\ \ j< h;\ \ j{+}{=}s\ \ $}
\STATE $substituteImg = bImg$
\STATE $substituteImg[i{-}k{:}i{+}k, j{-}k{:}j{+}k] = aImg[i{-}k{:}i{+}k, j{-}k{:}j{+}k]$
\STATE $[y_1', y_2', ..., y_c'] = CNN(substituteImg)$
\STATE $bMap[i, j] = y_{bId}' - y_{bId}$ \commentclr{// decrease of benign class probability}
\STATE $aMap[i, j] = y_{aId} - y_{aId}'$ \commentclr{// increase of adversarial class prob.}
\ENDFOR
\ENDFOR
\end{algorithmic}
\end{algorithm} 

\begin{figure}[t]
 \centering
 \includegraphics[width=0.9\columnwidth]{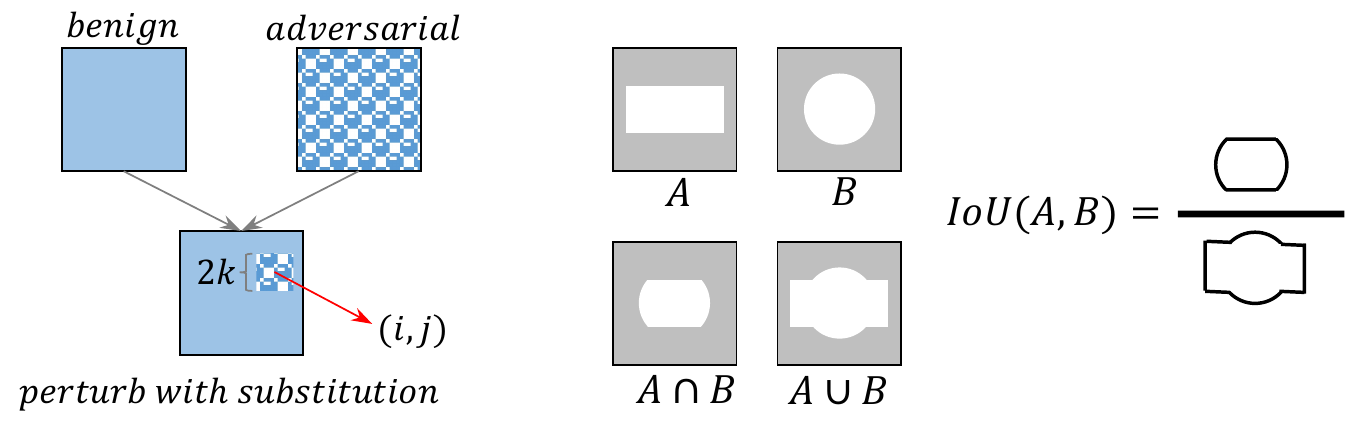}
 \caption{(Left) Image substitution/perturbation to probe the image's vulnerability. (Right) Union, intersection, and IoU computation.}
 \label{fig:iou}
\end{figure}

\textit{\textbf{Image-Perturbation Measure.}} Using a single forward pass of the CNN, we can generate the RFs of all neurons (of the last convolutional layer), obtaining a rough idea of which regions each neuron cares about. The next question is which regions are more vulnerable to the adversarial attack. By connecting these more vulnerable regions and neurons that care about such regions, we can uncover the vulnerable neurons. We quantify the vulnerability of different image regions by introducing a \textit{vulnerability map}, which is generated as follows (see \autoref{alg:vulmap} and \autoref{fig:iou} left). For a small region of $2k{\times}2k$ pixels centered at pixel ($i$, $j$), we substitute this small region of the benign image with the corresponding region of the adversarial image (lines $7$--$8$), feed the new image into the CNN (line $9$), and record the probability change at pixel ($i$, $j$). We consider both the decrease of the benign class's probability and the increase of the adversarial class's probability (lines $10$--$11$). Extending this process to all pixels (lines $4$--$5$), we produce two vulnerability maps, i.e., the $bMap$ and $aMap$ in lines $10$--$11$, sharing the same size with the input. 
These two maps are then binarized by a threshold so that they are comparable with the RF of each neuron.
Next, we compute the intersection over union (IoU, \autoref{fig:iou} right) between each neuron's RF and one of the binarized vulnerability maps selected from $bMap$ or $aMap$.
Sorting the neurons by their corresponding IoU identifies the most vulnerable ones (relating to either decrease of benign or increase of adversarial class's probability) as their focused regions overlap more with the vulnerable regions. In essence, \textit{this measure relies on the perturbation (i.e., region substitution) of input images.}

\begin{figure}[tb]
 \centering
 \includegraphics[width=0.7\columnwidth]{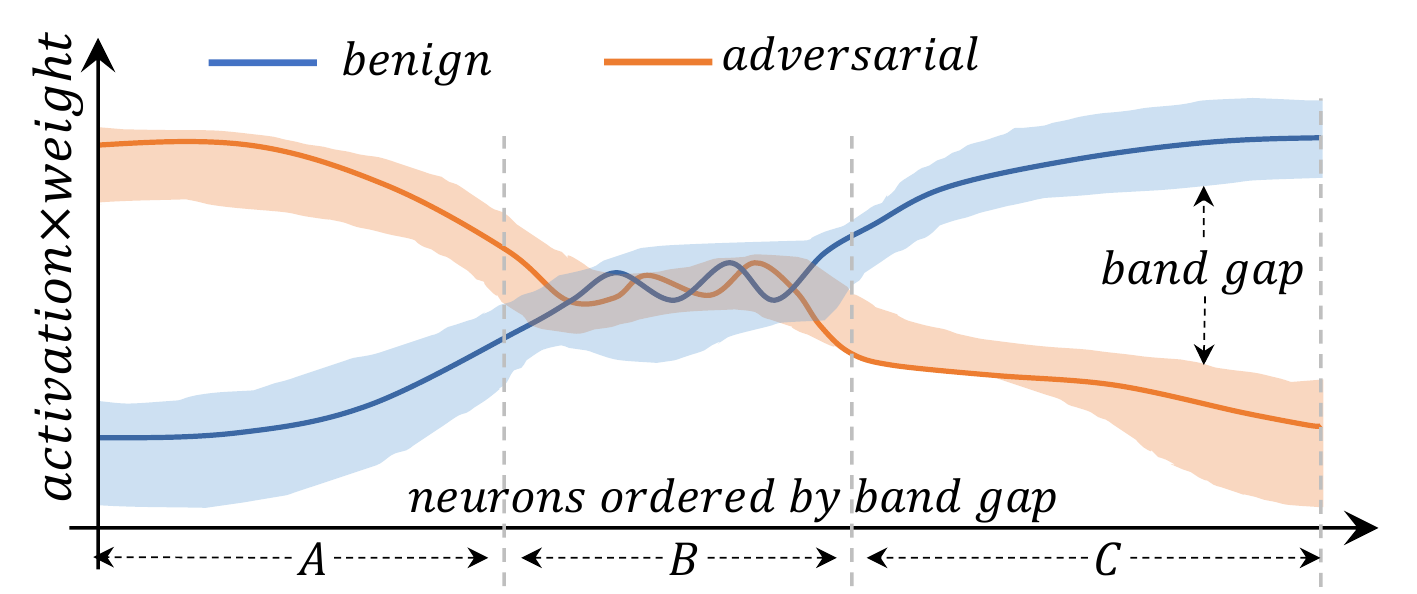}
 \caption{Using the band gaps to measure the vulnerability (i.e., the shift of the neurons' activation${\times}$weight) to adversarial attacks. The neurons are ranked by the magnitude of their band gaps. The values of band gaps are positive in Region A, zero in Region B, and negative in Region C. In the analysis, neurons to the left of Region A and to the right of Region C are of most interest.}
 \label{fig:shift}
\end{figure}

\textbf{\textit{Neuron-Perturbation Measure.}} 
The feature maps of the last convolutional layer of a CNN are aggregated through a pooling layer to convert them to an $n$D vector.
For example, in \autoref{fig:receptive}, the 512 feature maps of the size $14{\times}14$ are aggregated into 512 values (i.e., $n{=}512$).
Afterward, the dense layer combines the $n$D vector into a $c$D vector ($c$ is the number of classes) through a weighted sum to generate the logits for different classes. 
Note that the logits are values that produce class probabilities after applying normalization with a softmax function.
The $n$D vector, denoting the active level of the $n$ neurons (or called activation), is different for each input image but should share a roughly similar pattern across images within the same class. We, therefore, compute a confidence interval (CI) of the $n$D vectors corresponding to images in the same class. 
As shown in \autoref{fig:shift}, we use a curve to represent the $n$D vector of a benign image (each point on the curve corresponds to one neuron's activation) and a band to represent the CI of the class that the image belongs to. 
The same set of information is visualized for an adversarial counterpart. 
Similar to the \secondv{}, we further color the curves and bands based on the corresponding benign and adversarial labels.
We can consider that the adversarial attack shifts the curve of the benign images towards the trend of the adversarial band.
We can substitute the value of each benign neuron with the adversarial neuron to investigate the final probability change and then use this change to measure the neurons' vulnerability. Sorting with this measured vulnerability is equivalent to sorting the two curves or bands by the gap between them. Our explorations show that sorting by the gap between the two bands provides a more stable result compared with sorting by the difference between the two curves, i.e., more reflecting the activation changes at a class level.
The detailed computation of a band gap ($BG$) is formulated as:
    \begin{equation*}
        BG\left(\left(CL_a, CU_a\right), \left(CL_b, CU_b\right)\right) = \begin{cases}
                                            CU_a - CL_b &\text{if $CL_b > CU_a$} \\
                                            CL_a - CU_b &\text{if $CL_a > CU_b$} \\
                                            0 &\text{otherwise}
                                        \end{cases}
    \end{equation*}
where $(CL_a, CU_a)$ and $(CL_b, CU_b)$ stand for the lower and upper bounds of the CIs for the adversarial and benign classes, respectively.
As shown in \autoref{fig:shift} where the above sorting is applied, neurons in a region $A$/$C$ are excited/inhibited by the adversarial attack. Neurons in a region $B$ are not significantly affected by the attack. This view is horizontally zoomable to handle an excessive number of neurons.

We consider another fact that even when some neuron has a very high activation, the weight from the dense layer for the corresponding neuron could be minimal.
In this case, after the weight and activation are multiplied to produce the logits, the highly activated neuron's contribution to the prediction could be small. 
Therefore, to capture the actual impact on the prediction, we use the curve, together with the band, to represent the multiplication result of the $n$D vector and the corresponding weight from the dense layer (i.e., activation${\times}$weight), instead of the $n$D vector. Note that the benign and adversarial classes have different weights. For example, \autoref{fig:receptive} shows the structure of a 10-class CNN and the dense layer has the weight of a size $512{\times}10$ (10 rows). Each row of this weight matrix corresponds to the weight of one class.

\textit{\textbf{Visualization.}} The \thirdv{} and the \fourthv{} are designed to implement the above solutions.
The \thirdv{} (\autoref{fig:system}-c) serves as the starting point of the neuron space exploration and shows the plot explained in \autoref{fig:shift}. 
When hovering over any $x$-position (corresponding to one neuron) in the plot, a pop-up tooltip is shown with detailed information about the neuron. Users can sort the neurons through the different introduced measures (i.e., the image-perturbation measure based on either $bMap$ or $aMap$ and the neuron-perturbation measure). These measures help users expose the most vulnerable neurons and select one of their interests for further exploration.

The \fourthv{} (\autoref{fig:system}-d) presents two types of information to convey the vulnerability of the currently selected image from the \secondv{} together with the semantics of an image feature captured by the neuron selected from the \thirdv{}.
The first type of information is the selected neuron's RFs for the benign (\autoref{fig:system}-d1) and adversarial (\autoref{fig:system}-d2) images.
The two RFs come along with the predicted probability distribution on the right of each RF.
The examples in \autoref{fig:system}-d1 and \autoref{fig:system}-d2 have the probabilities of 1 for \texttt{cat} and \texttt{dog}, respectively; however, for the case where several classes have a non-zero probability, the view shows multiple bars.
The second type of information is the two vulnerability maps: one highlights regions where the image perturbation decreases the benign class's probability the most (\autoref{fig:system}-d3), and the other highlights regions where the image perturbation increases the adversarial class's probability the most (\autoref{fig:system}-d4).
While we show the top-20\% vulnerable pixels by default, this can be interactively changed with a slider placed in the view.

\subsubsection{How to Explore an Excessive Number of Neurons?}
Although the neurons can be sorted by the specified vulnerability measure (\autoref{sec:newmetrics}),
it is still inefficient to inspect, for example, hundreds of neurons one by one.
In fact, after users explored the semantics of a neuron, a natural question for the next step is if and what neurons behave similarly. 
Once we identify a group of similar neurons, by only examining one representative neuron from the group, we may understand the behaviors of the entire group, making the exploration more efficient.

We address this question by hierarchically clustering~\cite{mullner2011modern} the neurons and visualizing the hierarchy with an interactive dendrogram, as shown in the \fifthv{} (\autoref{fig:system}-e). As a dissimilarity measure used for clustering, we use the $L_2$ distance between two RFs to measure the corresponding neurons' dissimilarity.
We want to note that the IoU between two RFs or the $L_2$ distance between the corresponding activation maps could be used instead; however, based on our comparisons included in our Supplementary Material, there is no obvious advantage of one measure against the others. 
When users select a neuron from the \thirdv{}, its location in the dendrogram is highlighted, as shown in the red line at the far right of \autoref{fig:system}-e. 
From this location, users can easily identify neurons with similar behavior as similar neurons share the same ancestors and are generally located close to each other in the dendrogram.
Also, to see the RFs summarized across similar neurons, the user can click multiple nodes from the dendrogram, which selects the clicked nodes and their descendants.
For example, in \autoref{fig:system}-e, the root node of the subtree annotated with the dashed line is selected and so the corresponding descendants are highlighted, as shown in \autoref{fig:system}-e1.
For all selected neurons, the system computes the union of their RFs and updates the \fourthv{}, accordingly.
Users can also switch the \textit{union} operation to \textit{intersection} through a toggle switch at the top of the view.

This grouping of neurons based on their RFs' similarities and hierarchical clustering is a new approach to providing a summary of neurons' behaviors. 
By collectively using the vulnerability measure-based ordering described in \autoref{sec:newmetrics} and this grouping, we can uncover neurons that are potentially influential on the prediction but difficult to identify when only relying on the vulnerability measure-based ordering.
For example, the most vulnerable neuron and neurons that have behavior similar to it would not be placed closely in the \thirdv{} because of their activation${\times}$weight differences. 
However, when adversarial attacks aim at the most vulnerable neuron, these similar neurons are also likely influenced by the attacks. 
Even for the above case, users can first select the most vulnerable neuron from the \thirdv{} and then find similar neurons using the \fifthv{}.
Thus, our grouping approach not only enables efficient neuron space exploration but also supports a more comprehensive exploration of the neuron space.

\subsection{Cross-space Exploration}
\label{sec:cross-space}
The image space carries the semantics of different classes, while the mechanisms of how a CNN works on the images reside in the neuron space. Domain experts frequently need to jointly analyze both spaces to relate the semantics with the CNN's working mechanisms and interpret adversarial attacks (\textbf{R3}). We exemplify how our coordinated visual analytics system supports cross-space analysis through the three scenarios listed below.

\textit{\textbf{One Image--One Neuron.}} 
This scenario explores relations between one selected image and one selected neuron to interpret the neuron's behavior on a specific image. 
Specifically, after users select one image in the \secondv{} and one neuron in the \thirdv{}, the neuron's RF is extracted and overlaid onto the selected image. In \autoref{fig:system}-d1, we can clearly see that the neuron captures the face region of the \texttt{cat} image. In contrast, the neuron can no longer extract the \texttt{cat}'s face from the counterpart adversarial image, revealing the functionality change of the neuron caused by the adversarial perturbation.

\textit{\textbf{One Neuron--Multiple Images.}} The semantics captured by a neuron from a single image may not be able to sufficiently explain the neuron's behavior. This second exploration scenario allows users to select multiple images, and use the captured semantics from the selected images to gain more comprehensive interpretations of the neuron's behavior.
For example, when clicking the RF of the benign image in the \fourthv{}, a pop-up window lists the top-$m$ images that either maximally excite the neuron (sorting option 1) or have the highest prediction confidence (sorting option 2).
Each of the images is shown with its RFs to explain what the neuron has captured. 
On the other hand, when performing the same interaction on the RF of the adversarial image, the system first extracts the top-$m$ images for the benign counterpart using the same procedure above and then shows a pop-up window listing the extracted benign images' adversarial counterparts together with RFs.
These listed images work as a context to provide more visual evidence, helping users build their confidence when interpreting the functionality of a neuron. For example, the six images shown in the top left of \autoref{fig:case_study1} indicate that most of the RFs capture the \texttt{cat}'s face region, consolidating the behavior of the neuron. 

\textit{\textbf{One Image--Multiple Neurons.}} The large volume of neurons raises the need to explore them concurrently. The \fifthv{} uses the RFs of neurons on a selected image to cluster the neurons and allows the exploration of RFs in groups. For example, the selection of the dendrogram node in \autoref{fig:system}-e1 extracts all neurons belonging to this tree branch. The union or intersection of their RFs on the benign and adversarial image is visualized in the \fourthv{}. 
When applying the intersection, this exploration reveals the common behavior across the neurons.
On the other hand, the union of the RFs shows these neurons' behavior as a whole.

\section{Case Studies}
\label{sec:casestudy}
We first use a relatively small but well-known image dataset, CIFAR10~\cite{cifar10}, to demonstrate the usage of our system. The demonstration specifically presents how we answer the two focused research questions using our approach. Next, we use a larger dataset, Places365~\cite{zhou2017places}, to verify the generalizability of our solutions.
While we select specific classes in each case for the illustration, attacks on other classes can be investigated and interpreted in a similar manner.

\textbf{Adversarial Attack Settings.} 
For the PGD attack algorithm, we adopted the implementation by Madry et al.~\cite{madry2018towards}. This algorithm has three hyper-parameters: the number of iterative steps $\kappa$, the bounding neighborhood's radius (or the perturbation budget) $\epsilon$, and the iteration step size $\alpha$. 
For the attacks on both the CIFAR10 and Places365 datasets, we set the parameters to be: $\kappa{=}7$, $\epsilon{=}8/255{=}0.0314$, and $\alpha{=}2/255{=}0.00784$.

\subsection{CIFAR10 Dataset}
The CIFAR10 dataset~\cite{cifar10} contains 10,000 32x32 RGB images composing 10 classes in the test set.
Our studied CNN is a pretrained ResNet18~\cite{he2016resnet}, which correctly predicts 8,200 out of the 10,000 images.
We then perturb these true positive images with PGD attacks. All images are successfully attacked and plugged into our system for analysis.

\subsubsection{Categorizing Adversarial Images}
First, we focus on the \firstvScatterSingle{} of adversarial images, colored by the true labels or the adversarial predictions.
As shown in the \firstv{} in \autoref{fig:system}, the adversarial images with the same prediction outcomes tend to form large clusters (\autoref{fig:system}-a2),
whereas small clusters of the images with the same true labels exist within each large cluster (\autoref{fig:system}-a3).
This pattern indicates that while adversarial images with the same prediction are perturbed to contain similar features (i.e., features of the predicted class),
the images with the same true label also have their own attacked patterns.

\subsubsection{Understanding Attacks with the Image-Perturbation Measure}
\label{sec:cat-dog_case}

During the above image-space exploration using the \firstvScatter{}, we notice that there is a large proportion of \texttt{cat} images (the red cluster) being perturbed and misclassified into \texttt{dog} images (brown), 
as annotated in~\autoref{fig:system}-a2 and a3.
We then select this cluster to review their details in the \secondv{}.

As shown in \autoref{fig:system}-b, we first sort the images by the increasing order of their perturbation magnitudes; consequently, images perturbed less by the attacks are arranged at the upper side of the view. 
For these images, even with small perturbations, the adversarial attacks achieve drastic changes in the resulting probabilities, as shown by the bars above and the values below each image pair.
For example, while most of the benign images shown in \autoref{fig:system}-b have close to $100$\% probability to be predicted as \texttt{cat}, most of their adversarial counterparts have close to $100$\% probability to be \texttt{dog}.
As annotated with the green rectangle in \autoref{fig:system}-b, 
we select the pair with the smallest perturbation for further investigation. 

The \thirdv{} is updated based on this selection, as shown in \autoref{fig:system}-c.
To reveal vulnerable neurons along with their extracted image features, we sort the neurons by the image-perturbation measure using the vulnerability map (i.e., $bMap$ in \autoref{alg:vulmap}).
Then, we zoom into the top 25 neurons, as annotated with the dashed line in \autoref{fig:system}-c.
Afterward, we select one neuron of interest, neuron 403.
We are interested in this neuron as it still has only a small gap in activation${\times}$weight between the benign and adversarial images and, consequently, the neuron-perturbation measure cannot reveal this type of successfully attacked neurons. 

We further examine the corresponding RFs and vulnerability maps in the \fourthv{} (\autoref{fig:system}-d).
We first observe that the vulnerability maps for both benign-probability decrease (\autoref{fig:system}-d3) and adversarial-probability increase (\autoref{fig:system}-d4) contain a large proportion of the \texttt{cat}'s face region. 
This indicates that perturbations around the face region are one reason for the misclassification of this \texttt{cat} image as a \texttt{dog} image.
In \autoref{fig:system}-d3,  we see another focus around the feet region, which can be considered as another reason for the probability decrease of the \texttt{cat} class.
We also inspect the RFs of the chosen neuron, as shown in \autoref{fig:system}-d1 and d2.
Even though this neuron has similar activation${\times}$weight for both benign and adversarial images, the RFs are totally different (benign: face, adversarial: feet region). 
This difference in the RFs provides interpretations of how the adversarial attacks alter neuron 403's behavior (i.e., the focus region changes from the face to feet).

\begin{figure}[t]
 \centering
 \includegraphics[width=0.8\columnwidth]{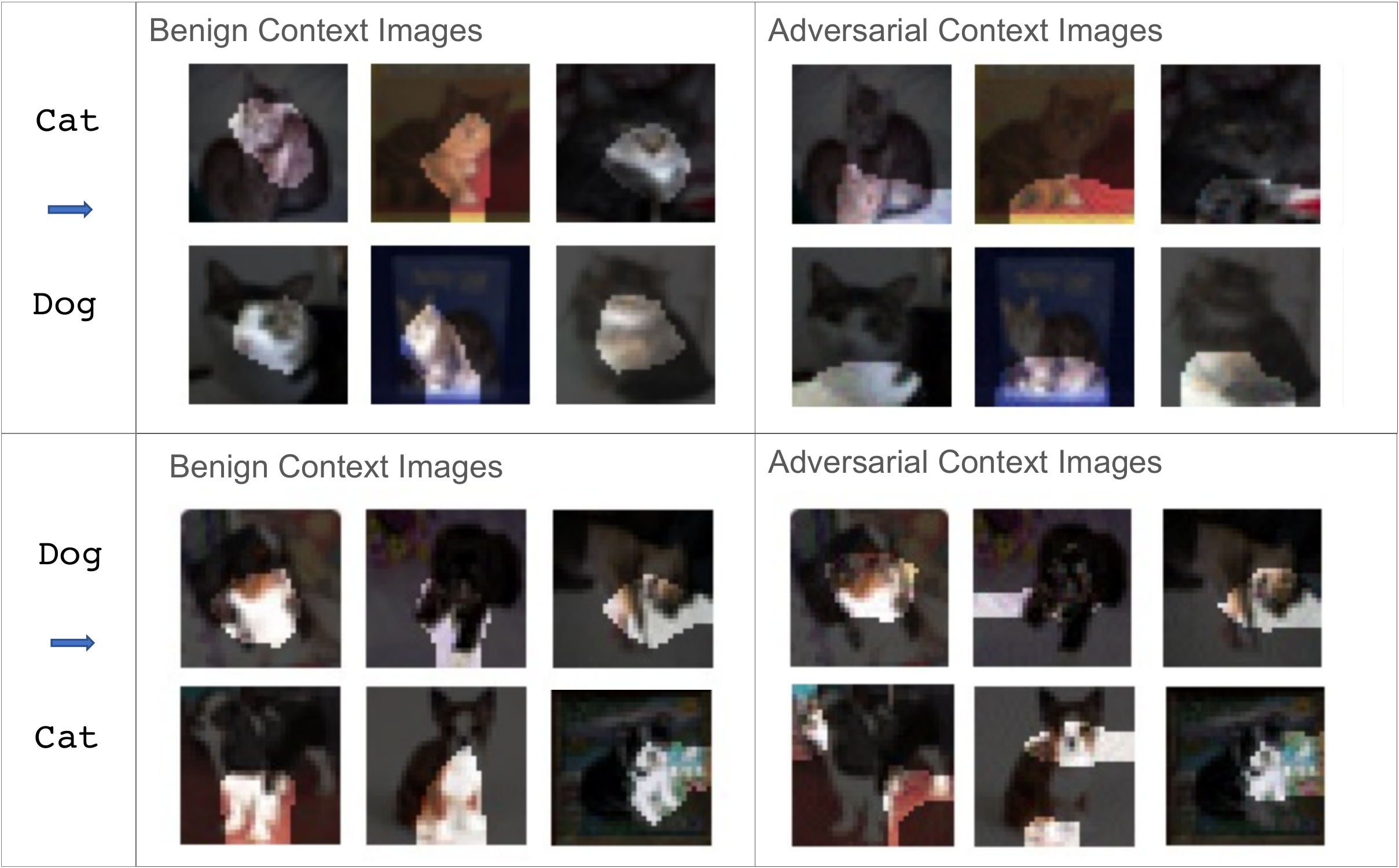}
 \caption{Context images for the \texttt{cat} v.s. \texttt{dog} case. Among the set of \texttt{cat} images misclassified as \texttt{dog} (top), cats' faces are captured in most of the benign images (top left), whereas cats' feet or lower bodies are captured in most of the adversarial counterparts (top right). This indicates that the same neuron behaves differently before and after the adversarial attacks.
 The behavior change can also be seen in \texttt{dog} images misclassified as \texttt{cat} (bottom), where the benign and adversarial images capture dogs' feet/lower bodies and faces, respectively.
 }
 \label{fig:case_study1}
\end{figure}

To verify this interpretation, we click on each RF to check the context images, which are also \texttt{cat} images perturbed to be misclassified as \texttt{dog}.
As presented in \autoref{fig:case_study1} (top), neuron 403 indeed tries to capture the face region for benign images but the feet region for adversarial images.
This result can be interpreted that the perturbations around the face region disturb neuron 403's recognition of the \texttt{cat} ``face'' feature, while the perturbations around the feet region lead neuron 403 to focus on the ``feet'' feature, which can be easily misunderstood as the \texttt{dog}'s.
To further validate this interpretation, we select \texttt{dog} images that are attacked to be \texttt{cat} from the \firstvMatrix{} and visualize the context images, as shown in \autoref{fig:case_study1} (bottom).
When feeding the benign images of \texttt{dog}, neuron 403 captures the feet region, while its RF shifts to the face region for the adversarial images.
This provides the inference that neuron 403's role is to capture the face region for \texttt{cat} images but the feet region for \texttt{dog} images, 
and the adversarial attacks try to mislead neuron 403 to capture the adversarial label's feature by weakening the benign label's feature as well as enhancing the adversarial label's feature.

\begin{figure*}[tb]
 \centering
 \includegraphics[width=0.8\textwidth]{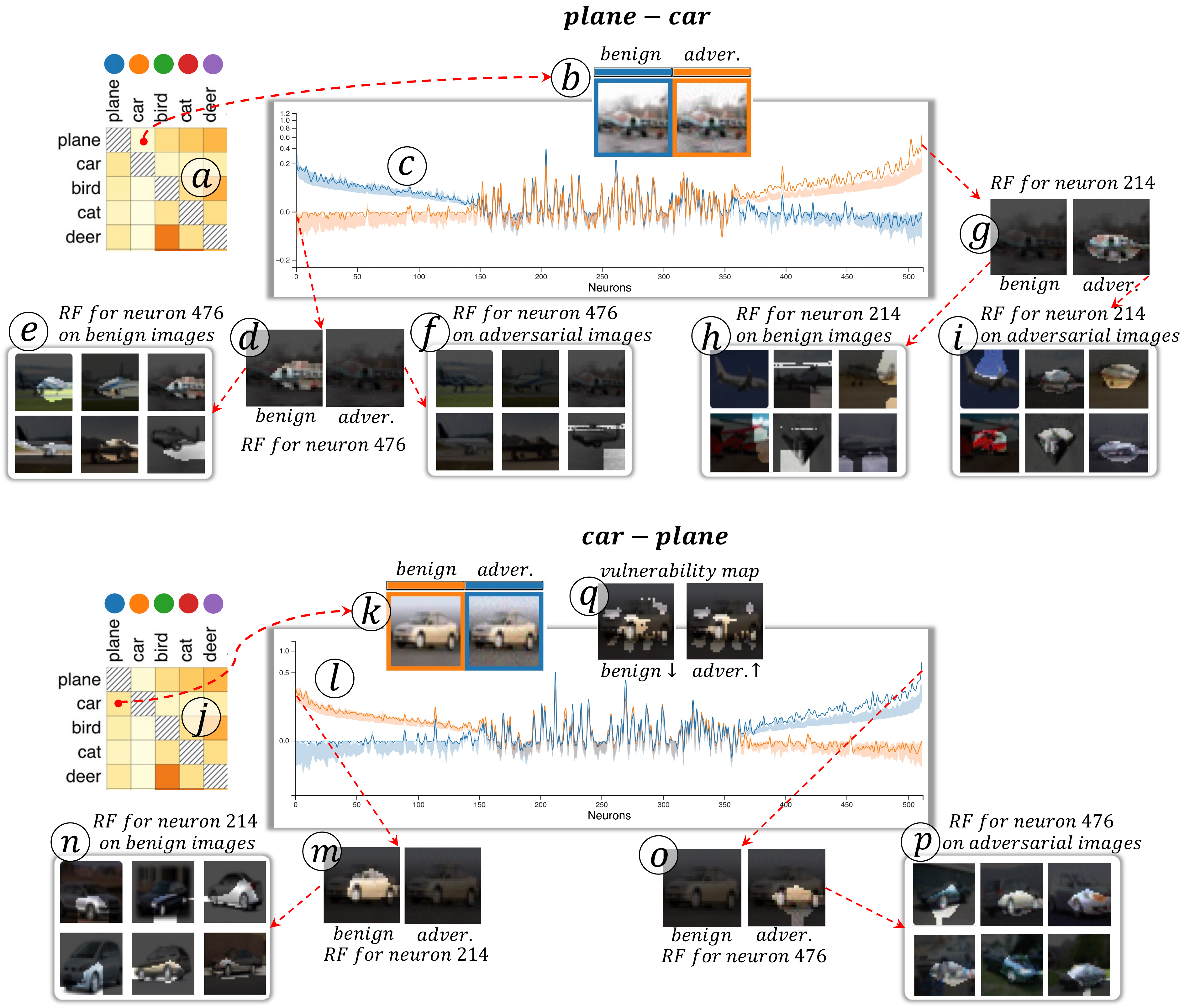}
 \caption{The \texttt{plane}-\texttt{car} case started by selecting the cell from the \firstvMatrix{} (a). The \secondv{} lists the corresponding images, such as those in (b).  The CNN uses neurons 476 and 214 located at both ends of $x$-axis of the \thirdv{} (c) to identify the wheel regions for \texttt{plane} and \texttt{car} classes, respectively (d-e, m-n). The adversarial attacks fool the CNN into recognizing a \texttt{plane} as a \texttt{car} by (f) disabling neuron 476 but (g--i) exciting neuron 214. 
 The roles of neurons 476 and 214 can be further examined by reviewing the related visualizations to the \texttt{car}-\texttt{plane} cell (j--p). 
 From these observations, we can infer that the CNN still extracts wheel features from a \texttt{plane} image, but it believes the wheels are from a \texttt{car} as they are extracted by neuron 214.}
 \label{fig:casecar}
\end{figure*}

\subsubsection{Understanding Attacks with the Neuron-Perturbation Measure}
\label{sec:neuron-perturb-cifar}

We next analyze the case where the attacks cause confusion between  \texttt{plane} and \texttt{car} classes using the neuron-perturbation measure.
We first select the \texttt{plane}-\texttt{car} cell (\autoref{fig:casecar}-a) from the \firstvMatrix{} to retrieve \texttt{plane} images (benign label) that are mispredicted into \texttt{car} (adversarial label) after adversarial perturbations. In the \secondv{}, sorting the images by the change in prediction probabilities, we identify images that the CNN yields very different predictions. One of such image pairs is shown in \autoref{fig:casecar}-b. 
Based on the filled blue and orange bars in \autoref{fig:casecar}-b, the CNN's prediction for this image is changed from \texttt{plane} to \texttt{car} with strong confidence.
However, it is hard to visually differentiate the benign and adversarial images.

Next, selecting this pair of images, we explore the corresponding neuron space. Sorting the neurons by the ``Gap Size'' (i.e., using the same sort showed in \autoref{fig:shift}), we easily identify the most vulnerable neurons on the two sides of \autoref{fig:casecar}-c. For example, neuron 476 is the left-most one, which is active when processing the benign image (blue curve) or images of \texttt{plane} (blue band); however, it becomes inactive (close to 0, $y$-axis value) or negatively active (i.e., decreasing the benign class's probability) when processing the adversarial image (orange curve) or images of the \texttt{car} class (orange band). The change of this neuron decreases the benign class's probability the most. Checking the details of this neuron on the selected image with the RFs, as shown in \autoref{fig:casecar}-d, we notice that this neuron extracts the wheel and tail features of the \texttt{plane} for the benign image but extracts nothing after the adversarial attacks. Clicking the RFs brings a group of representative images and their RFs (six images are shown here), in which the wheel regions of \texttt{plane}s are consistently extracted in the benign images (\autoref{fig:casecar}-e), but almost nothing is extracted from the adversarial counterparts (\autoref{fig:casecar}-f). Thus, the adversarial attacks seem to successfully stop this neuron from working properly.

While knowing this detail is helpful to interpret the attacks, we still need to find out whether or not the adversarial attacks prevent the CNN from being aware of the wheel feature. As the next step, we further explore several neurons that increase the adversarial probability the most on the right of \autoref{fig:casecar}-c, e.g., neuron 214 (the right-most one). This neuron is inactive when processing the benign \texttt{plane} image but is active when processing the adversarial image. As shown in \autoref{fig:casecar}-g, its RF on the selected adversarial image shows the extraction of the wheel region of the \texttt{plane}. Similar observations are also found from the context images (\autoref{fig:casecar}-h and i), confirming that the CNN still extracts the wheel regions after the attacks but through a different neuron.

Afterward, by selecting the \texttt{car}-\texttt{plane} cell (\autoref{fig:casecar}-j), we switch the analysis to \texttt{car} images mispredicted into \texttt{plane} and choose an image pair with a significant probability shift (\autoref{fig:casecar}-k). 
\autoref{fig:casecar}-l shows the \thirdv{} sorted by ``Gap Size'', where the neurons are reversely ordered when compared to those in \autoref{fig:casecar}-c.
The left-most neuron is 214, which contributes the most to the decrease of the benign class's probability (\texttt{car}). As the RF shows in \autoref{fig:casecar}-m, this neuron extracts the wheel of a \texttt{car} on benign images (supported by the context images in \autoref{fig:casecar}-n as well). However, the adversarial image successfully prevents the neuron from extracting this feature. The right-most neuron in \autoref{fig:casecar}-l is 476; it captures nothing from \texttt{car} images but extracts the wheel features from perturbed images (\autoref{fig:casecar}-o and p). 

\textit{\textbf{Insight.}}
In short, the CNN can extract wheel regions from both the \texttt{car} and \texttt{plane} classes but through very specific neurons, e.g., in our case, neuron 214 for the \texttt{car} and neuron 476 for the \texttt{plane} classes, respectively. 
The adversarial attacks fool the CNN into misrecognizing a \texttt{plane} image as a \texttt{car} image by causing the following situation with the perturbation: (1) the neurons that extract the wheel from \texttt{plane} images can extract nothing, (2) while the neurons that extract the wheel from \texttt{car} images can extract the wheel feature. The CNN then believes the image is a \texttt{car} as it recognizes the wheels as \texttt{car}'s. This also indicates that the wheel feature is a non-robust feature when differentiating \texttt{car} and \texttt{plane}. During the exploration, we also checked the vulnerability maps for different selected images. For example, \autoref{fig:casecar}-q shows those for the \texttt{car} image. These vulnerability maps helped us verify that the wheel regions of the image are indeed vulnerable to attacks. 

\subsubsection{Exploring Neurons with a Hierarchy}

We here present our explorations with neurons related to the \texttt{cat} image shown in \autoref{fig:system}-d and the \texttt{plane} image shown in \autoref{fig:casecar}-b to demonstrate the effectiveness of the \fifthv{} to efficiently explore a large number of neurons. 512 neurons for the \texttt{cat} image are explored in a bottom-up order, whereas 512 neurons for the \texttt{plane} image are explored in a top-down order.  

\begin{figure}[tb]
 \centering
 \includegraphics[width=\columnwidth]{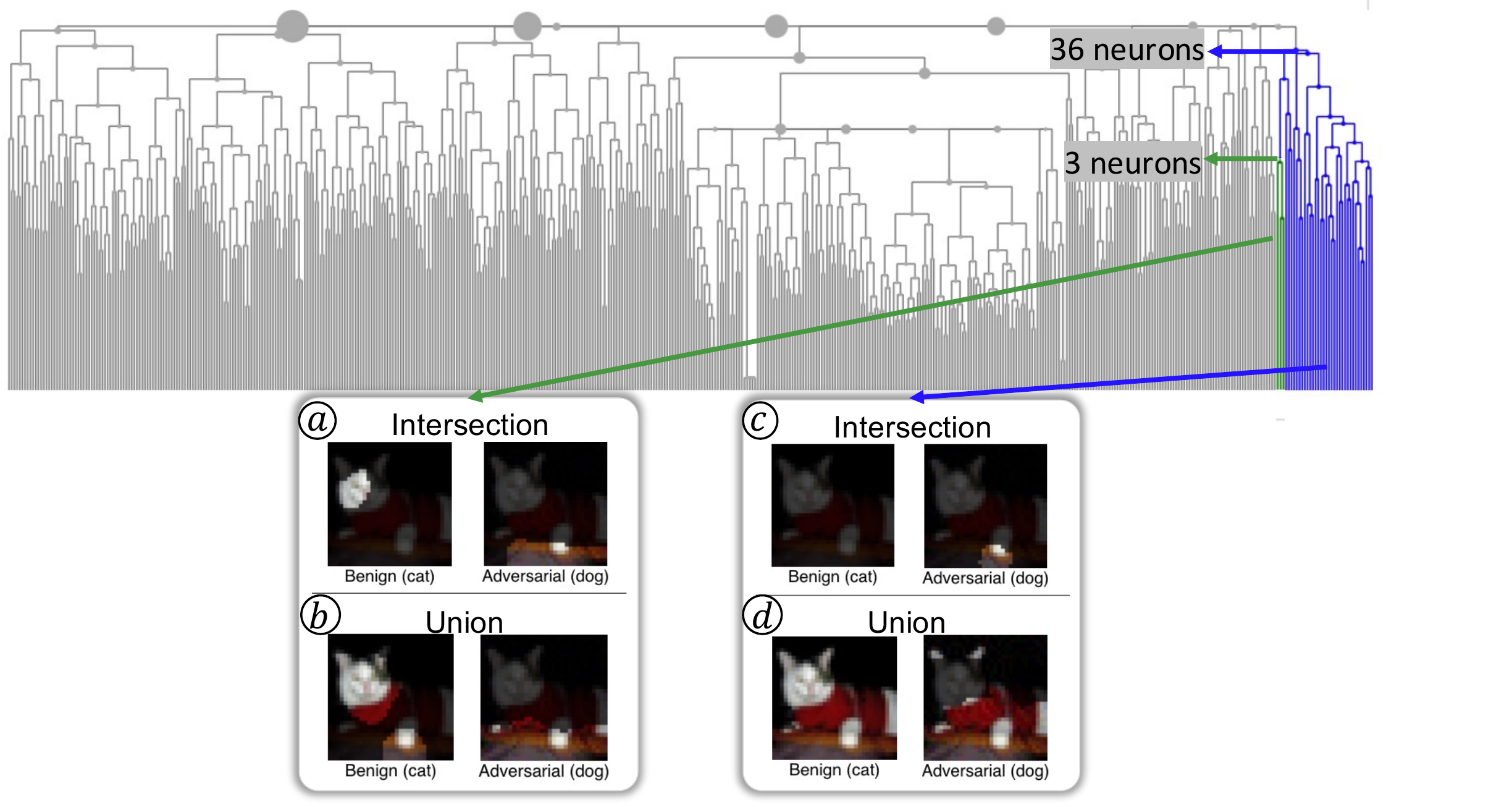}
 \caption{Bottom-up exploration of the \texttt{cat} image using the \fifthv{}. As the size of a cluster increases (i.e., adding more neurons in a cluster), the image features captured by the neurons are more varied in the benign images (e.g., while the 3-neuron union highlights the face and foot, the 36-neuron union shows the entire body). However, in the adversarial image, the captured image feature stays similar (e.g., both the 3- and 36-neuron unions do not highlight the cat's face).}
 \label{fig:hierarchical_explore}
\end{figure}

\textbf{\textit{Bottom-Up Exploration.}}
In~\autoref{sec:cat-dog_case}, neuron 403 is inspected and interpreted.
In the \fifthv{}, this neuron is highlighted with red, as seen at the right side of \autoref{fig:system}-e.
Here, the dendrogram is generated based on the L2 distance between each neuron's benign and adversarial RFs.
To quickly identify neurons that behave similarly to neuron 403, we locate clusters where the highlighted neuron belongs.

As shown in \autoref{fig:hierarchical_explore}(top), we select the small green cluster consisting of 3 neurons and examine the intersection and union of RFs.
Through the intersection (\autoref{fig:hierarchical_explore}-a), we notice that all 3 neurons capture the \texttt{cat}'s face region in the benign image and the \texttt{cat}'s feet region in the adversarial image.
In~\autoref{fig:hierarchical_explore}-b, the union of RFs indicates that the neurons' behavior varies a little in the benign image, but every neuron captures the feet region in the adversarial image.

We move two levels up and select the bigger blue cluster, which contains 36 neurons.
This time, the union of RFs covers almost the entire region of the image (\autoref{fig:hierarchical_explore}-d), whereas the intersection of RFs covers nothing of the image (\autoref{fig:hierarchical_explore}-c).
This indicates that the behavior of neurons on the benign image has more variation within the cluster.
However, for the adversarial image, while the intersection of RFs still captures the feet region, the union of RFs covers a wider area (mostly on the lower half of the body), in which the face region is still not covered.
This indicates that all neurons within the cluster try to capture the feet region but ignore the face region.
Therefore, although these neurons may capture different features of the benign image, they are altered to capture similar features of the adversarial image.

From these two different cluster selections, we observe that, indeed, there exist neurons that behave similarly with others.
One interesting insight is that the adversarial attacks perturb the images so that many neurons (36 neurons) act to make similar mistakes (i.e., overlooking \texttt{cat}'s face and focusing on their feet).

\begin{figure}[tb]
 \centering
 \includegraphics[width=\columnwidth]{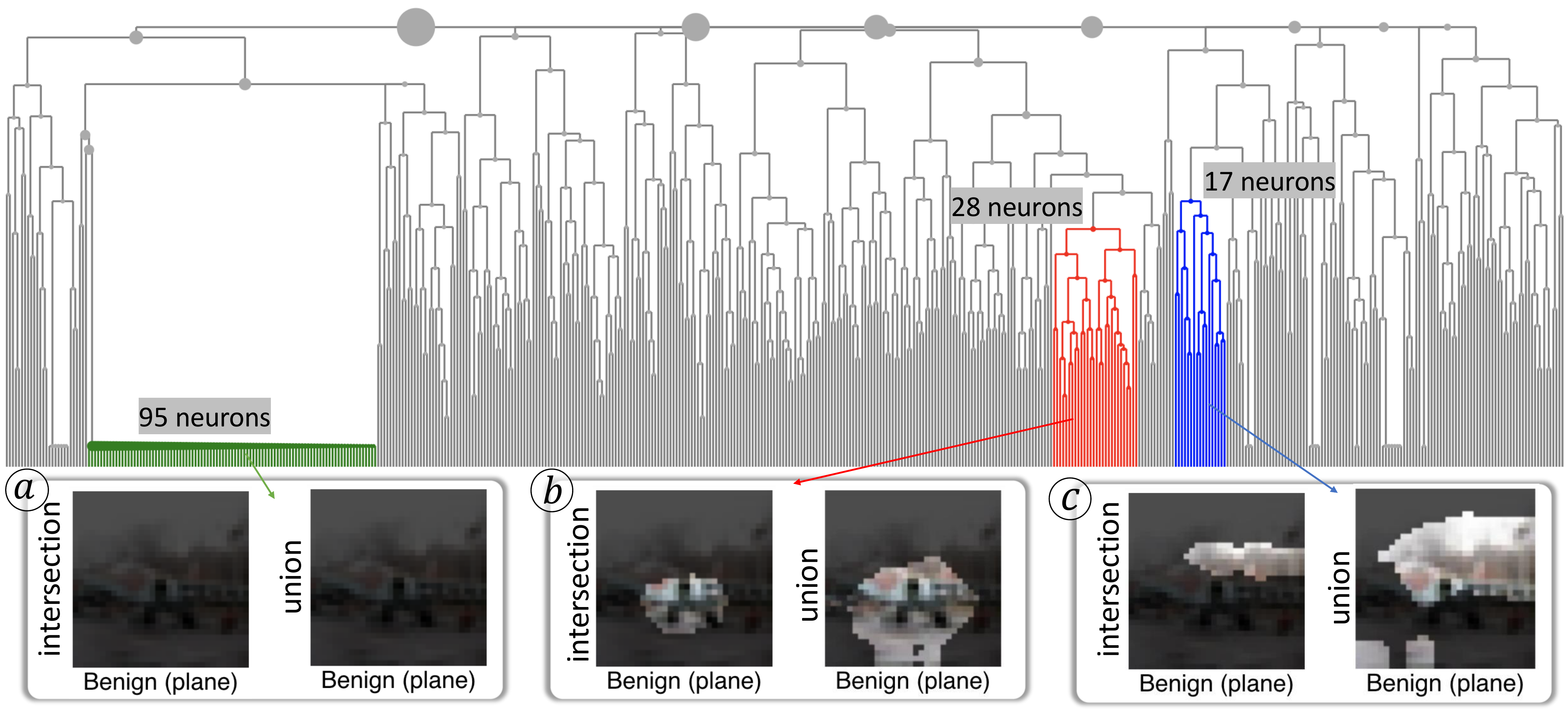}
 \caption{Top-down exploration of the \texttt{plane} image using the \fifthv{}. Through this exploration, the distribution of features is roughly summarized. For example, the green-cluster neurons (a) capture no \texttt{plane} features, and thus do not need further investigation. The red cluster (b) captures mostly the body and wheel regions of the \texttt{plane}, while the blue cluster (c) captures the sky right above the \texttt{plane}.}
 \label{fig:dendrogram}
\end{figure}

\textit{\textbf{Top-Down Exploration.}} \autoref{fig:dendrogram} shows the exploration of the neurons for the benign \texttt{plane} image. From the dendrogram, the big green cluster immediately draws our attention. This cluster has 95 neurons and all of them capture very similar image features, as indicated by the very low vertical positions of the tree nodes. Selecting this cluster and exploring the intersection and union of the corresponding 95 RFs (\autoref{fig:dendrogram}-a), we see that this cluster includes all the ``dead'' neurons, i.e., they all capture nothing from the \texttt{plane} image. These neurons can be excluded from further analysis. We also explore several other big clusters to explore their correspondence to the image features. For example, the red cluster of 28 neurons all capture the body and wheel region of the \texttt{plane}, as shown by the intersection of their RFs (\autoref{fig:dendrogram}-b). Meanwhile, individual neurons also extract some surrounding features, as shown by the union of their RFs. For the blue cluster of 17 neurons, the common region is the sky above the \texttt{plane} (\autoref{fig:dendrogram}-c). Some neurons also extract the lower-left corner of the image, making the functionalities of neurons in this cluster a little more diverse, which is indicated by the higher vertical positions of the tree nodes. Even when only exploring the big clusters, we can already observe an overall picture of how image features have been distributed across the neurons. For any clusters, users can interactively traverse the dendrogram from a parent to its child to narrow down the analysis focus.

\subsection{Places365 Dataset}

We now demonstrate an analysis case with the Places365 dataset~\cite{zhou2017places}.
We use the validation set as the input,
which contains 100 images for each of the 365 scene categories.
The CNN is a pretrained wide ResNet18 model~\cite{zagoruykowide2016}.
19,745 out of 36,500 images are correctly predicted before the attacks, while the attacks successfully cause the misclassification of 12,298 images out of the 19,475 images.
The number of classes (i.e., 365) is much larger than the CIFAR10 dataset.
Therefore, we first demonstrate neuron space exploration through example images directly selected in the backend. Then, we present two more complete explorations within a subset of images.

\subsubsection{Image Example from Backend Selection}

\begin{figure}[tb]
 \centering
 \includegraphics[width=0.85\columnwidth]{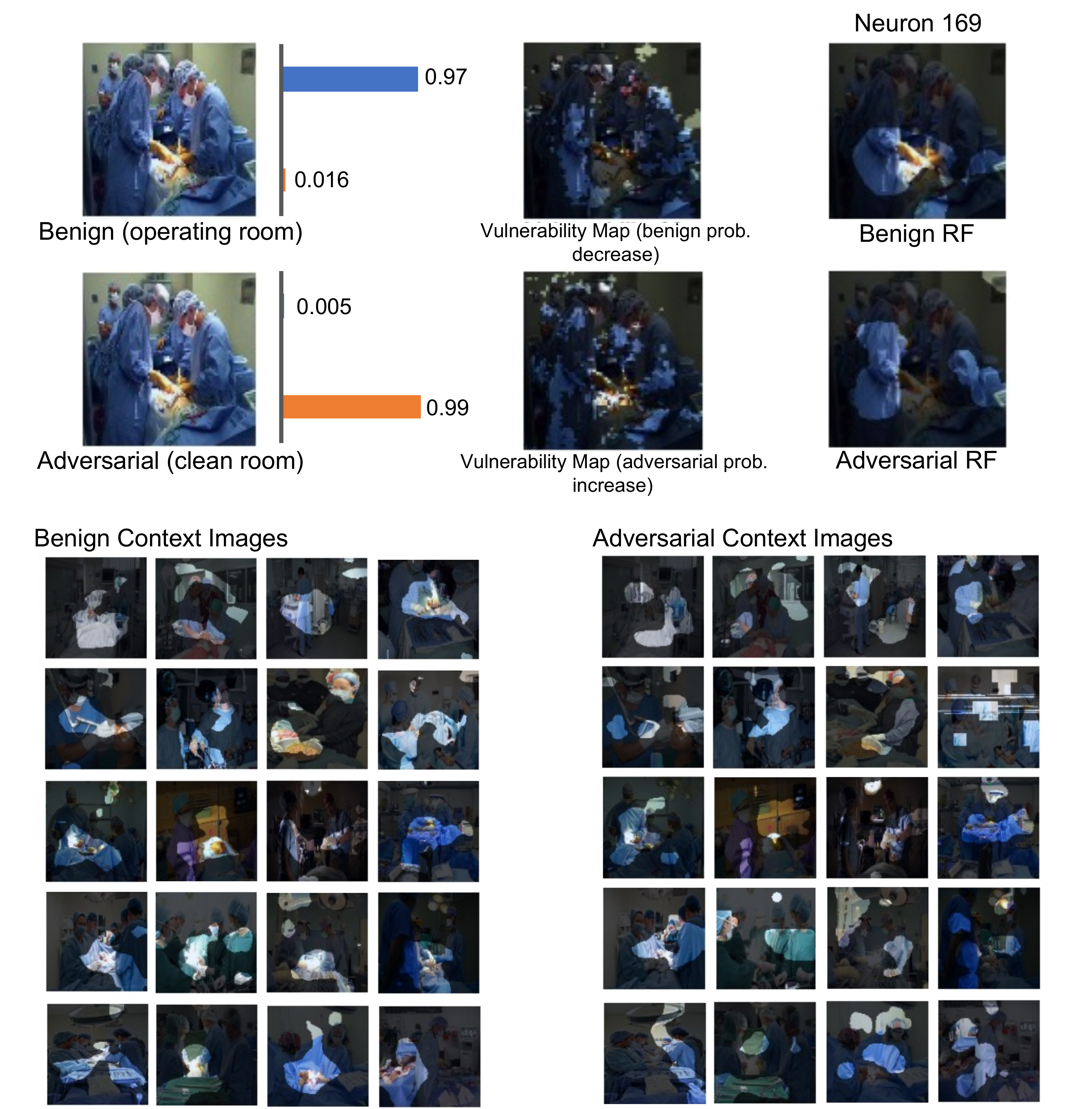}
 \caption{Demonstration of the case study with the Places365 dataset. The selected neuron mostly captures the operating table and surgeons' hands in the benign image but the surgical blue or green gowns in the adversarial image. 
 This finding can be verified through the context images.}
 \label{fig:case_study_places365}
\end{figure}

We select an \texttt{operating room} class image that is misclassified as \texttt{clean room} after the attacks, as shown at the top-left corner of \autoref{fig:case_study_places365}.
We select this image because it shows the largest change in the prediction probabilities when compared with other attacked images.
In addition, as a large number of \texttt{operating room} images are misclassified as \texttt{clean room},
it is meaningful to analyze the attacking patterns of this group of images.

As shown in \autoref{fig:case_study_places365},
although there are few or no observable differences between the benign and adversarial images,
the CNN's probability outputs are significantly different.
We examine the vulnerability maps presented on the top middle of \autoref{fig:case_study_places365}.
Through the highlighted area,
we observe that perturbations around the surgical gowns decrease the benign probability the most,
whereas perturbations around the operating table and hands region increase the adversarial probability the most.

Based on the \textit{image-perturbation measure}, we then select neuron 169 for further exploration, which is one of the top neurons with the largest overlap between the RF and the vulnerability map for the benign probability decrease.
From the RFs at the top right of \autoref{fig:case_study_places365}, we can see a clear difference in the RFs of benign and adversarial images:
the neuron focuses more on the hands and operating table regions in the benign image
but focuses more on the surgical gown regions in the adversarial image.

We then examine the RFs of the benign and adversarial context images, which are listed as grids in \autoref{fig:case_study_places365} (bottom).
Similar to the selected image, most benign context images highlight regions around the hands and operating table,
while most adversarial context images highlight regions around the surgical gowns.

\textit{\textbf{Insight.}}
It can be concluded that the perturbations inhibit the \texttt{operating room} feature existing around the operating table, which is important to distinguish \texttt{operating room} and \texttt{clean room} images.
At the same time, the perturbations around the surgical gowns make neuron 169 shift its focus to the gown feature, which can be expected to be also seen in \texttt{clean room} images.
In this way, the adversarial attacks succeed in fooling the CNN.

\subsubsection{Explorations in a Subset of Images}
\label{sec:plces365_subset}

\begin{figure*}[tb]
 \centering
 \includegraphics[width=\columnwidth]{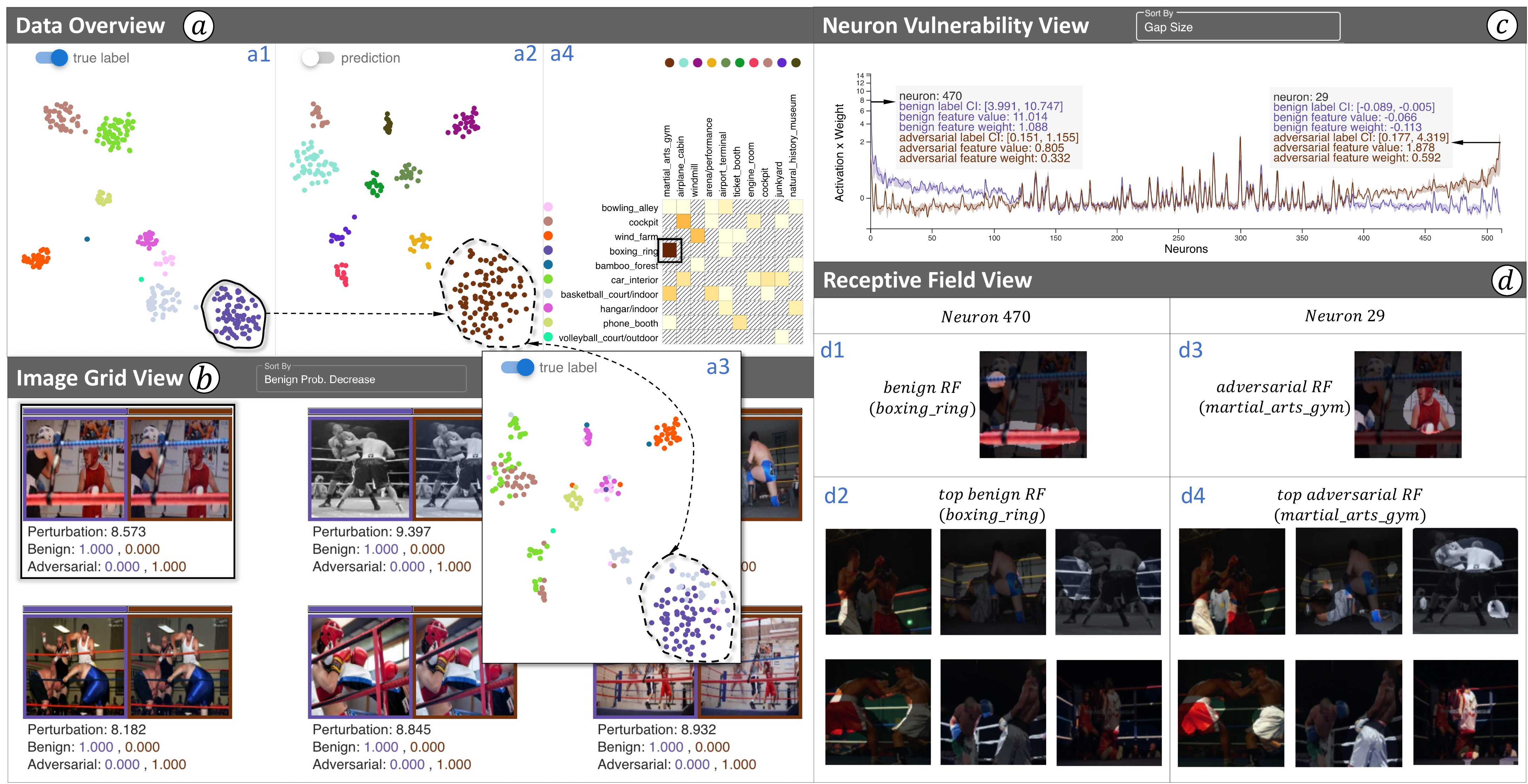}
 \caption{Investigation of the adversarial attacks on a variety of Places365 images.
 Here we analyze only a subset of classes (see a), where a large number of images are misclassified after the attacks. From a4, we select the \texttt{boxing\_ring}--\texttt{martial\_arts\_gym} cell which  contains the largest number of misclassified images.
 Then, we use the neuron-perturbation measures for further examination, as presented in b, c, and d. Through identifying the most vulnerable neurons from c and observing the RFs in d, we see that the attacks seem to hide the rope region uniquely seen in \texttt{boxing\_ring} but expose the human-related feature that can also be seen in \texttt{martial\_arts\_gym}.}
 \label{fig:case1_places365}
\end{figure*}

To fully utilize our visual analytics system, we first narrow down our analysis target to a manageable size of classes (e.g., ten classes).
This is needed mainly because of the visual scalability of the \firstvMatrix{} (i.e., 365 rows/columns are difficult to be displayed) and the color encoding (i.e., humans cannot distinguish too many different colors).
From the 365 classes in Places365, we select the top-10 \textit{benign classes} that contain the largest number of misclassified images after the attacks (e.g., \texttt{bowling\_alley} and \texttt{cockpit} rows in \autoref{fig:case1_places365}-a4). 
Then, as \textit{adversarial class} pairs, we further select the classes corresponding to the top-10 adversarial labels observed within the selected benign classes (e.g., \texttt{martial\_arts\_gym} and \texttt{airplane\_cabin} columns in \autoref{fig:case1_places365}-a4).

As shown in \autoref{fig:case1_places365}-a1 and a2, we see that both benign (a1) and adversarial (a2) images are clearly grouped by their labels.
From \autoref{fig:case1_places365}-a3, which applies the color encoding based on true labels to a2, we also see that many adversarial classes contain a few benign classes.
For example, the \texttt{martial\_arts\_gym} cluster annotated by the dashed line mainly consists of \texttt{boxing\_ring}, \texttt{basketball\_court/indoor}. 
This pattern can also be observed in the \firstvMatrix{} shown in \autoref{fig:case1_places365}-a4.

For specific cells of interest selected from the \firstvMatrix{}, we present two cases explaining adversarial attacks through the neuron-perturbation measure and the image-perturbation measure, respectively, as in \autoref{sec:cat-dog_case} and \autoref{sec:neuron-perturb-cifar} for the CIFAR10 dataset.

\textit{\textbf{Case 1: Understanding Attacks with the Neuron-Perturbation Measure.}}
We first reason about the attacks that fool the CNN into mispredicting \texttt{boxing\_ring} images as the \texttt{martial\_arts\_gym}, which take the largest portion in the subset, as indicated by the dark orange color in \autoref{fig:case1_places365}-a4.
After retrieving the corresponding images, we sort the images by the decreasing order of their benign class probability drops in the~\secondv{} (\autoref{fig:case1_places365}-b).
The images at the top show drastic changes in their prediction probabilities after adversarial attacks, where the CNN is tricked to have zero probability for the benign class but 100\% probability for the adversarial class.
We select the first pair of images to explore the neuron space.

In the neuron space, the attacks can be explained through comparisons between RFs of the two most vulnerable neurons identified with the neuron-perturbation measure. 
After sorting the neurons by ``Gap Size'' in the \thirdv{}, the most vulnerable neurons are identified on both ends of the $x$-axis (\autoref{fig:case1_places365}-c).
Neuron 470 located on the left end is largely activated by the benign image but also largely suppressed by the adversarial image. In contrast, neuron 29 on the other end shows the reverse active pattern.
Therefore, we compare neuron 470's RF of the benign image against neuron 29's RF of the adversarial image as each RF represents the regions that mostly activate each neuron.
As shown in \autoref{fig:case1_places365}-d1 and d3, we discover that neuron 470 captures the ropes surrounding the \texttt{boxing\_ring}, whereas neuron 29 captures the humans inside the ring.
This difference is further validated through the comparison of RFs on multiple \texttt{boxing\_ring} images and their adversarial counterparts (\autoref{fig:case1_places365}-d2, d4).

\textit{\textbf{Insight.}} In conclusion, the ropes surrounding the \texttt{boxing\_ring} is an essential feature for the CNN to identify the \texttt{boxing\_ring}, while humans within the \texttt{boxing\_ring} is a confusing feature that highly contributes to the \texttt{martial\_arts\_gym} prediction.
The adversarial attacks seem to use perturbation to hide the rope feature and to amplify the human-related feature, leading to fooling the CNN into misclassification. 

\begin{figure*}[tb]
 \centering
 \includegraphics[width=0.84\columnwidth]{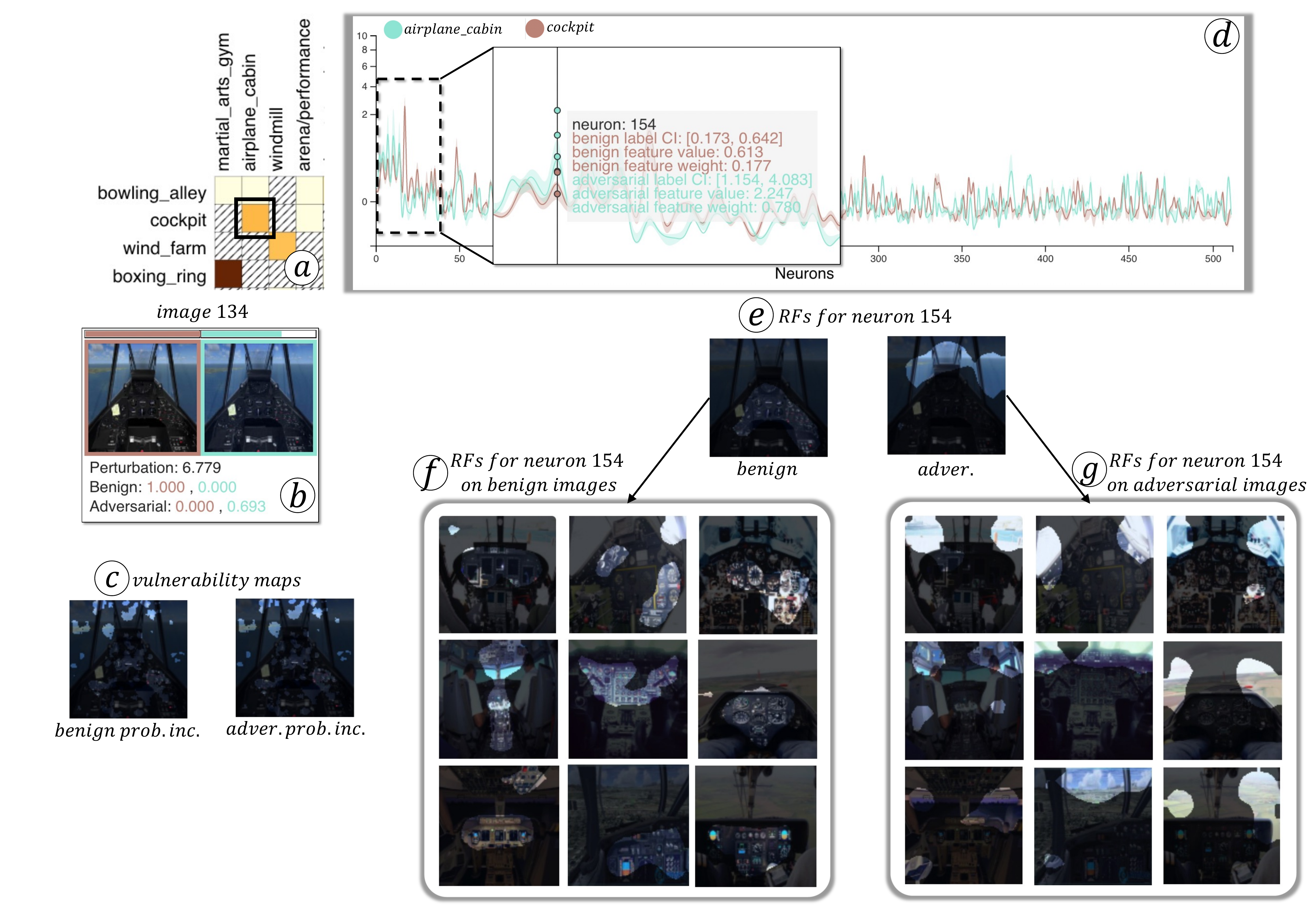}
 \caption{Reasoning adversarial attacks through the image-perturbation measure. 
 We investigate the attacks on images corresponding to the \texttt{cockpit}--\texttt{airplane\_cabin} cell (a).
 Similar to the other cases, we visualize the related information (b--g). Neuron 154 focuses on different features in the benign and adversarial images (e, f, and g), which demonstrates how the attacks shift the functionality of the neuron.}
 \label{fig:case2_places365}
\end{figure*}

\textit{\textbf{Case 2: Understanding Attacks with the Image-Perturbation Measure.}}
We here present another case explaining the attacks corresponding to the \texttt{cockpit}--\texttt{airplane\_cabin} cell, as highlighted in \autoref{fig:case2_places365}-a. 
After sorting the related images by the decreasing order of their benign class probability drops, we select the top image (image 134 in \autoref{fig:case2_places365}-b), where the adversarial class probability increases from 0.000 to 0.693 by the perturbation.
The selected image's vulnerability maps (\autoref{fig:case2_places365}-c) indicate that the most vulnerable regions are around the window and instrument panel.

We further explain the attacks related to the selected cell by exploring the neuron space. 
In the \thirdv{} (\autoref{fig:case2_places365}-d), neurons are sorted by their overlaps between RFs and the vulnerability maps.
We then select neuron 154 as it has the highest IoU score among the top neurons and it is activated by both benign and adversarial images.
As shown in \autoref{fig:case2_places365}-e, the RFs of neuron 154 focuses on the instrument panel and window regions in the benign and adversarial images, respectively.
\autoref{fig:case2_places365}-f and g, which list the RFs for other \texttt{cockpit} images misclassified as \texttt{airplane\_cabin}, show similar patterns to the above, validating our findings.

\textit{\textbf{Insight.}} The adversarial perturbation successfully hides the unique feature of \texttt{cockpit} (instrumental panel), while amplifying the common feature shared by both \texttt{cockpit} and \texttt{airplane\_cabin} (window). As the functionalities of each neuron are entangled, after the perturbations, a neuron capturing the instrumental panel feature may shift its focus onto the window region, which decreases the \texttt{cockpit} probability and increases the \texttt{airplane\_cabin} probability.

\section{Domain Experts' Feedback}
\label{sec:feedback}
We conducted an evaluation of the system with two domain experts ($E_1$ and $E_2$) working on various types of adversarial attacks, e.g., adversarial examples, data poisoning, and adversarial code analysis. Both are full-time research scientists from a research lab and hold a Ph.D. degree in computer science (with a focus on adversarial deep learning). The study lasted for ${\sim}90$ minutes for each expert, in which we explained our two targeted research questions, guided the experts to explore the cases we explained in \autoref{sec:casestudy}, and paused at the findings that triggered the experts for comments. Lastly, we conducted think-aloud discussions and open-ended interviews with the experts to collect their feedback. Neither expert co-authored this paper (for objective evaluations).

In general, both experts provided positive feedback on the way we answer the two research questions and the interactivity of our coordinated visual analytics system. $E_1$ supported our choice of focusing only on the last convolutional layer of a CNN for analysis as the high-level features extracted by neurons of this layer provide the human-understandable semantics for interpretation. Also, our vulnerability map was very inspiring to him after seeing the results in \autoref{fig:casecar}. The observation motivated him to fine-tune the CNN by masking the wheel regions (i.e., the vulnerable pixels) and enforcing the CNN to learn other \texttt{car}-related features to improve the model's robustness. This idea is along the same lines as the recent work of Ilyas et al.~\cite{ilyas2019adversarial}. $E_2$ enjoyed the coordination between different views of our system. He commented that their colleagues work more with different Python code snippets to generate visual evidence for interpretations. However, the discrete visual reasoning process significantly hindered their analysis efficiency. Our integrated and coordinated visualizations (e.g., cross-space explorations) provided a more friendly interface to involve humans in the analysis process. During the study, $E_2$ also had many intriguing thoughts triggered by our visual interpretation results. For example, he suspected the non-robust features between two classes (e.g., wheels for \texttt{car} and \texttt{plane} in \autoref{fig:casecar}) could be robust features between another two classes (e.g., \texttt{car} and \texttt{cat}). He also suggested pasting the wheel features from a \texttt{car} image to a \texttt{plane} image to probe the CNN's behavior. Furthermore, he seconded that our hierarchical exploration of neurons helps experts efficiently discover the ``dead" neurons or neurons with similar functionalities, which could be used to prune CNNs and compress them.

There are also several limitations and potential improvements suggested by the experts. First, $E_1$ commented that PGD does not guarantee to attack continuous image features and suggested considering the synergetic effects of multiple image features when perturbing an image to generate its vulnerability maps. Second, $E_2$ believed that different attacking methods (e.g., FGSM~\cite{goodfellow2015explaining}) might result in different vulnerability levels of the neurons. But, he also speculated that the neurons' vulnerability should not change too much when attacking the same CNN. This is a good task for us to validate and compare different adversarial attack methods in the future. Third, $E_2$ also suggested extending our work to other adversarial attacks, e.g., the attacks on graph data or source code, where (unlike images) the data instances may not carry much semantics. How to visually distill human-understandable semantics and use them to interpret adversarial attacks on other types of data is an interesting future direction to explore. 
\section{Discussion, Limitations, and Future Work}
\label{sec:discussion}

\textbf{\textit{Limitations.}} 
As pointed out by the experts, our current approach perturbs one image patch at a time when generating the vulnerability map (\autoref{fig:iou}), and does not consider the combined effects of multiple patches (as there are so many possible combinations). 
This is a common limitation of similar saliency map generation algorithms (e.g.,~\cite{zhou2015object}). 
A potential remedy for this issue is to randomly sample patch combinations during the perturbation process and evenly attribute the probability changes to the sampled patches. 
Secondly, the scalability issue of our current system is caused when analyzing the Places365 dataset. 
Specifically, the views for the image space exploration (\autoref{fig:system}-a) are not capable of handling a large number of classes (e.g., 100 classes). 
In practice, however, as the exploration usually focuses on a small number of classes, users can select their classes of interest in advance and only load the corresponding information into the system, as in our analysis demonstrated in \autoref{sec:plces365_subset}.
This selection would also help narrow down the number of data instances to be explored.
Lastly, our current evaluation relies on expert users with sufficient knowledge of adversarial attacks. 
In the future, we intend to recruit more users with different expert levels to thoroughly evaluate the usability of different visualization components.

\textbf{\textit{Generalizability.}} We want to emphasize four points in terms of generalizability. 
First, our approaches are applicable to both targeted and untargeted adversarial attacks~\cite{papernot2016limitations}. 
In fact, based on the instances' distribution from the  \textit{Prediction Matrix View}, we can even infer whether the studied attack is targeted or not (i.e., the cells for the targeted class should have darker color). 
Second, our way of identifying the vulnerable neurons and their follow-up interpretations is agnostic to the attacking algorithms and their parameters. 
While PGD is analyzed in this work because of its popularity and effectiveness, our system does not have any attacking-specific features and can be applied to other attacks, such as FGSM~\cite{goodfellow2015explaining}.
In addition, our methods can be adapted to analyze attacks that use the same algorithm but apply different parameters determining the attacking strengths, which is further described in the Supplementary Material.
Third, our visual analytics system is agnostic to the computation of RFs. Although the approximation method described in~\autoref{sec:receptive_field} is adopted for efficiency, it can be replaced with perturbation-based RF computations. The Supplementary Material provides a detailed comparison between different RF computations (e.g., the discrepancy map introduced by Zhou et al.~\cite{zhou2015object}).
Fourth, our work only focuses on adversarial attacks on CNNs, but we believe our general approach of identifying critical neurons to an attack through perturbation is applicable to other types of data (e.g., texts or graphs) and deep learning models.

\textbf{\textit{System Interactivity.}} The most computationally expensive part of our system is the generation of the vulnerability map, as it requires perturbing all possible image patches with numerous forward-pass of a target CNN (lines 4-5 in \autoref{alg:vulmap}). The stride parameter (i.e., the variable $s$ in \autoref{alg:vulmap}) also influences the performance. 
However, with $s{=}1$ in the CIFAR10 dataset, the generation of the vulnerability map still only takes about 3 seconds when using CPUs (specifically, we tested with Intel i7-6900K).
To further improve computational efficiency, we also provide an implementation utilizing GPUs, and this version takes about 0.3 seconds when using a machine with a graphics card, Nvidia Titan Xp. 
For the Places365 dataset, the vulnerability map can be efficiently generated with precomputed logits of all the perturbed images, which corresponds to line 9 in~\autoref{alg:vulmap}. 
For 266 images (of the size $224{\times} 224$) that we  select for our analysis, the pregeneration involves $266 {\times} 224 {\times} 224 {=} 13,346,816$ forwardings and takes about 3 hours with an Nvidia Quadro RTX 8000 graphics card.
After this pregeneration, the vulnerability map can be produced within about 0.003 seconds.
Although the latency may vary across different machine configurations, it does not significantly limit our analysis.
This is because (1) we can expect that the focused image is not updated very often, so as the computation of its vulnerability map; (2) \autoref{alg:vulmap} can utilize parallel computing and the CNN's forward-pass time is significantly reduced when we use a GPU-enabled machine; (3) we can easily adjust the stride to balance the trade-off between accuracy and latency. For example, increasing the stride from 1 to 2 does not radically change the appearance of the resulting vulnerability map but can reduce the computation time down to $1/4$. 
As the interactivity of a visual analytics system is a crucial factor for maintaining users' mental maps during analysis, extensively profiling the latency of our system with different CNN models on different hardware platforms is also part of our planned future work.

\textbf{\textit{Significance.}} 
This work focuses on understanding adversarial attacks mainly through instance-based analyses.
Our approach effectively uncovers model vulnerabilities in neurons' behaviors of a selected pair of images. Since the neurons' behaviors on a certain image group can often be approximated by the behaviors on a representative instance of the group, the instance-based analysis can provide concrete, intuitive interpretations for the representative instance.
Furthermore, our approach supports analysis of specific neurons (e.g. extreme neurons with the largest band gap) and provides supplemental information for various multi-level analyses.
To explore the behavior of groups of neurons, we design the \fifthv{}, which allows the user to conveniently select similar neurons and review their group behavior through the intersections and unions of their RFs.
Also, we visualize the sorted context images in the \fourthv{} to support findings derived from the selected image.
We demonstrate the applicability and generalizability of the introduced analytic workflow with multiple case studies.
In the future, we plan to support the analysis of global patterns for neurons on a group of input images.

\section{Conclusion}
In this work, we introduce a visual analytics approach to understanding the adversarial attacks happening in CNNs. Our approach aims to answer two questions: which neurons are more vulnerable and how they have worked in the prediction process. By perturbing the input images or neuron activations, we decompose the consequence of adversarial attacks into individual neurons and introduce several measures to rank the neurons by their vulnerability. Hierarchical clustering is used to structurally explore the excessive number of neurons and reveal their behavior similarity in benign/adversarial images. Integrating all these technical innovations, we design a visual analytics system, empowering users to conveniently explore the input images and CNN neurons jointly, and use the semantics carried by input images and extracted by CNN neurons to interpret adversarial attacks. Through multiple case studies and domain experts' feedback, we validate the efficacy of our visual analytics approach.

\begin{acks}
This research is sponsored in part by the National Institute of Health through grants 1R01CA270454-01 and 1R01CA273058-01 and by the Knut and Alice Wallenberg Foundation through Grant KAW 2019.0024.
\end{acks}

\bibliographystyle{ACM-Reference-Format}
\bibliography{main}
\end{document}